\documentclass[10pt,twocolumn,letterpaper]{article}

\usepackage{subfiles}
\usepackage{cvpr}
\usepackage{times}
\usepackage{epsfig}
\usepackage{graphicx}
\usepackage{amsmath}
\usepackage{amssymb}
\usepackage{caption}
\usepackage{import}
\usepackage{pdfpages}
\usepackage{titling}
\usepackage{bibunits}
\usepackage{microtype}
\usepackage{float}

\usepackage{siunitx}

\usepackage{mathtools}
\usepackage{booktabs}
\usepackage{cite}
\usepackage{subcaption}
\usepackage{enumitem}
\pretitle{\begin{center}\Large \bf\selectfont}
\posttitle{\par\end{center}}

\preauthor{\large \lineskip .5em \begin{center}\begin{tabular}[t]{c}\selectfont}
	\postauthor{\vspace*{1pt}\end{tabular}\end{center}}

\date{}

\usepackage[pagebackref=false,breaklinks=true,letterpaper=true,colorlinks,bookmarks=false]{hyperref}

\cvprfinalcopy %

\def\cvprPaperID{****} %

\begin{document}

\title{Actor-Action Video Classification \\ CSC 249/449 Spring 2020 Challenge Report}

\author{
Jing Shi\thanks{CSC 249/449 Spring 2020 Teaching Assistants.} \hspace{4mm} Zhiheng Li\(^*\) \hspace{4mm} Haitian Zheng\(^*\)\hspace{4mm} Yihang Xu\(^*\) \hspace{4mm}Tianyou Xiao\(^*\) \\
Weitao Tan\(^*\) \hspace{4mm} Xiaoning Guo\(^*\) \hspace{4mm}Sizhe Li\(^*\) \hspace{4mm} Bin Yang \hspace{4mm}  Zhexin Xu \hspace{4mm} Ruitao Lin\\
Zhongkai Shangguan \hspace{4mm} Yue Zhao \hspace{4mm} Jingwen Wang \hspace{4mm} Rohan Sharma \\
Surya Iyer \hspace{4mm} Ajinkya Deshmukh \hspace{4mm} Raunak Mahalik \hspace{4mm} Srishti Singh \\
Jayant G Rohra \hspace{4mm} Yipeng Zhang \hspace{4mm} Tongyu Yang \hspace{4mm} Xuan Wen \\
Ethan Fahnestock \hspace{4mm}  Bryce Ikeda \hspace{4mm} Ian Lawson \hspace{4mm} Alan Finkelstein \hspace{4mm} Kehao Guo  \\
Richard Magnotti \hspace{4mm} Andrew Sexton \hspace{4mm} Jeet Ketan Thaker \hspace{4mm} Yiyang Su \hspace{4mm} Chenliang Xu \\
University of Rochester
}

\maketitle

\begin{abstract}
This technical report summarizes submissions and compiles from Actor-Action video classification challenge held as a final project in CSC 249/449 Machine Vision course (Spring 2020) at University of Rochester~\cite{course_page}.
\end{abstract}
 \vspace{-1mm}
\section{A2D Dataset}
While previously computer vision researchers focus on human-centric action recognition task, Xu \etal~\cite{Xu_2015_CVPR} firstly generalize the action understanding problem via considering actions performed by various actors (\ie, adult, dog, bird, \etc). The Actor-Action (A2D) dataset~\cite{Xu_2015_CVPR} is proposed to study this new problem, which contains 3782 videos annotated with segmentation masks labeled with actor-action (\ie, a certain actor performs a certain action) categories. Seven actor classes and eight action classes are considered in A2D dataset. Note that some actor-action pairs are invalid, \eg, ``adult-fly'', ``car-climb'', \etc.

\begin{figure}[t]
  \centering\includegraphics[width=0.71\columnwidth]{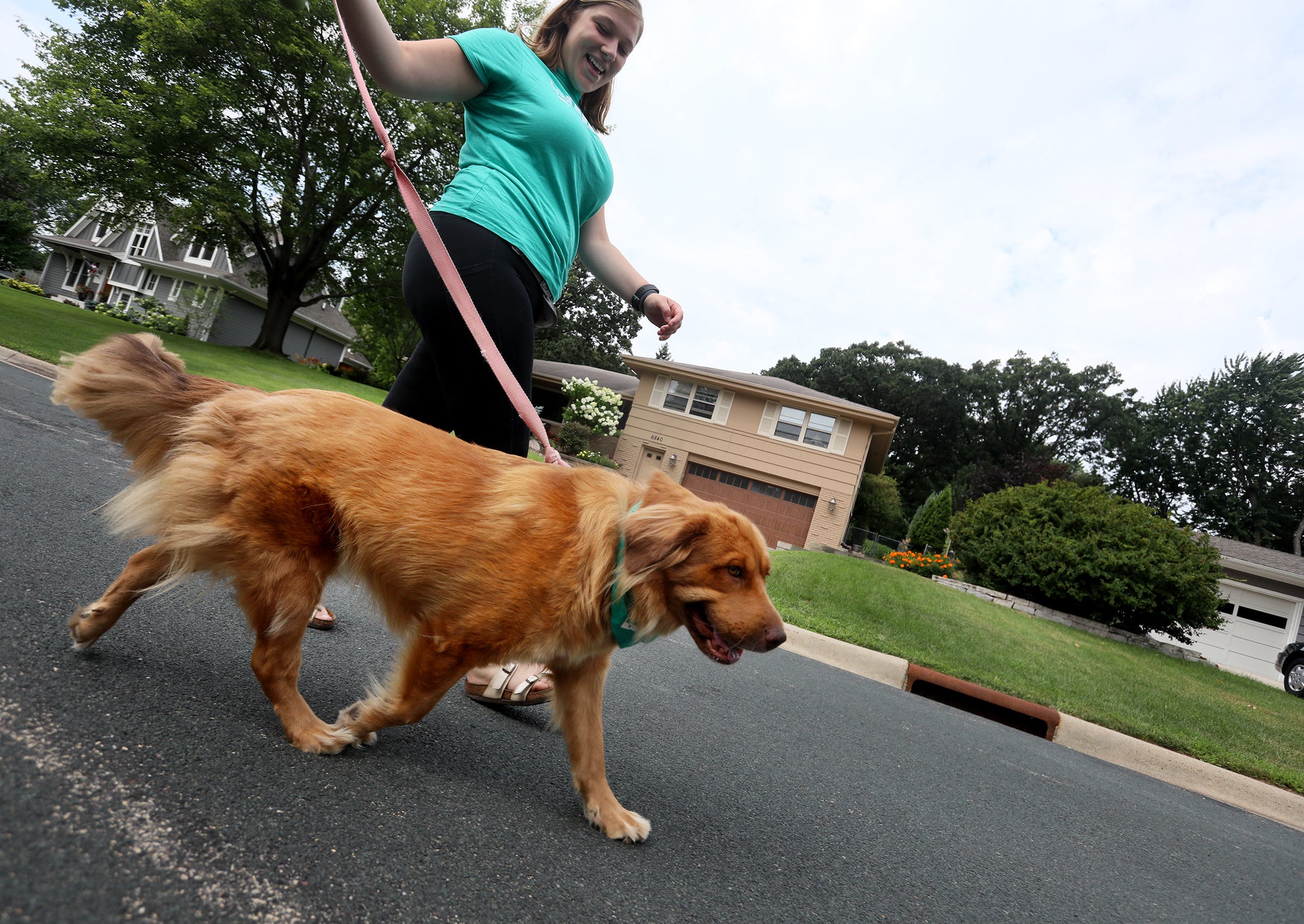}
  \caption{One example frame in A2D Dataset.}
  \label{fig:intro_teasing}
\end{figure}

\begin{table}[t]
	\centering
	\scalebox{0.9}{
	\begin{tabular}{c|ccc}
		\hline
		Group& Precision& Recall& F1-Score\\
		\hline \hline
		Alan&$\sim$41&$\sim$58&$\sim$46\\
		Ethan, Bryce and Ian & 42.0 & 59.3& 46.3\\
		Kehao & 32.5& 43.3& 34.6\\
		Richard, Andrew and Jeet & 15.6 & 36.7 & 19.1\\
		Rohan, Surya and Ajinkya & 54.0 & 67.0 & 57.0 \\
		Raunak, Srishti and Jayant & 50.6 & 57.6 & 51.6 \\
		Yiyang & N/A& N/A& N/A\\
		Yipeng, Tongyu and Xuan & 52.1 & 55.4 & 51.5 \\
		Zhongkai, Yue and Jingwen & 60.6 & 61.1 & 59.2 \\
		Bin, Zhexin and Ruitao & \bf{63.2} & \bf{66.8} & \bf{62.9}\\
		\hline
	\end{tabular}}
	\caption{The reported performances on the actor-action classification task of the A2D dataset~\cite{Xu_2015_CVPR}. The performances are measured in precision, recall and F1-score.}
    \label{tab:performance}

\end{table}

\section{Actor-Action Classification Challenge}
The actor-action classification challenge is to classify the key frames in the A2D videos into actor-action pairs.
More than one actor may occur in a frame, making it a multi-label classification task.
For example, the labels of Fig.~\ref{fig:intro_teasing} are ``adult-walking'' and ``dog-walking''.
Although this task is image level, the videos are available for temporal clues.

\section{Summary of Results}
Table~\ref{tab:performance} lists the actor-action classification performances of all the precipitated teams. The performances are measured in precision, recall and F1-score. From the table, the top three teams are Bin \etal, Zhongkai \etal and Rohan \etal, respectively. The best F1-score is 62.9. For more details of their methods, please refer to the paper in the following compilation.

{\small
\bibliographystyle{ieee_fullname}
\bibliography{egbib}
}
\clearpage

\newpage

\setcounter{section}{0}
\setcounter{figure}{0}
\setcounter{equation}{0}
\setcounter{table}{0}

\title{Multi-Label Actor-Action Classification with Per-Class Detection Methods}

\author{Bin Yang \hspace{4mm} Zhexin Xu \hspace{4mm} Ruitao Lin\\
Brain and Cognitive Sciences, University of Rochester\\
{\tt\small \{byang13, zxu53, rlin18\}@ur.rochester.edu}
}

\maketitle

\begin{abstract}
   We present multiple methods for multi-label actor-action classification based on the idea of per-class detection. We tactile this problem by converting multi-label classification into a series of independent object detection for every possible class. Specifically, we employ a ResNet backbone to extract base features from images, apply an attention mechanism to select the features, and then incorporate 43 parallel detectors to perform the task. We further utilize the temporal information in the A2D dataset to improve the network performance. Finally, we assemble three different networks together to achieve higher classification accuracy.
\end{abstract}

\begin{bibunit}[unsrt]
\section{Introduction}

To solve the multi-label classification problem in the A2D dataset \cite{xu2015can}, our basic idea is to detect the presence of each class independently. The simplest way to achieve this is training 43 models in parallel, each model for one possible class. Obviously, this is a computationally heavy method and there are a lot of redundancy in model parameters. To minimize the complexity of the model, we first make use of well-established CNNs to extract common features shared by all possible classes. We then build 43 independent detectors on top of the base feature map to predict the presence of each class.

We first implemented and trained three different models as described in following sections, then we combined them together to form an ensemble as our final working model.

\section{Per-class detection with ResNet (PCD-ResNet)}
\subsection{Architecture}

In this model, we first obtained a feature vector ($2048\times1$) from the last average-pooling layer of a ResNet-152 \cite{he2016deep}, and then built 43 detectors on top of it for each of the 43 classes. Each detector (43 detectors in total) is composed of a fully-connected layer ($2048\times2$) followed by a softmax operation, and gives the probability distribution (two probabilities in the range of 0 to 1) of the detection of one specific class. The output of the whole model is then a vector ($43\times1$) containing the detection probabilities of all 43 classes (Fig. 1).

\begin{figure}[t]
\begin{center}
   \includegraphics[width=1.0\linewidth]{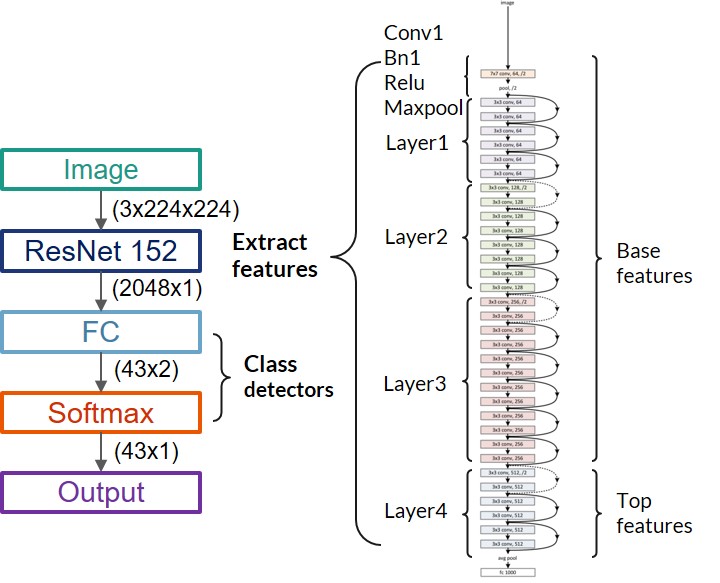}
\end{center}
   \caption{Architecture of PCD-ResNet. Left panel: the architecture of our PCD-ResNet. Right panel: structure of ResNet \cite{he2016deep}.}
\label{fig:1}
\label{fig:onecol}
\end{figure}

\subsection{Training parameters and model performance}

For the preparation of training dataset, we kept random rotation, random blur and random flip operations initially implemented in A2DDataloader, but removed the random crop operation. All these manipulations are meant to increase the variability of the training dataset hence to reduce the over fitting problem, however, we proposed that the random crop operation could actually be harmful to the training. The reason is that it reduced the number of positive training samples (reduced 26.8\% on training dataset and 28.1\% on validation dataset) for each class by cutting off actors from images (Fig. 2A). And in deed, the performance of the model dropped substantially when the random crop was applied (Fig. 2B, red versus magenta curves).

The model parameters were initialized with a ResNet-152 model pretrained on ImageNet \cite{imagenet_cvpr09}.
Log-loss was used for training:
\begin{equation}
    \mathcal{L}=-\sum_{i=0}^{N} {\big[y_i \log p_i + (1-y_i)\log (1-p_i)\big]}
\end{equation}
where $N=43$ is the number of classes, $y_i$ and $p_i$ are the ground truth and predicted label for the $i$-th class, respectively.

The optimization method here is stochastic gradient descent, with a learning rate $lr=\num{1e-5}$, momentum $=0.9$, and weight decay $=\num{4e-5}$ 

As shown in Fig. 2B (red line), the F1 score is close to 60 after 24 epochs, and then fluctuated below 60 and maximized to 59.4 at 70 epochs.

\begin{figure}[t]
\begin{center}
   \includegraphics[width=1\linewidth]{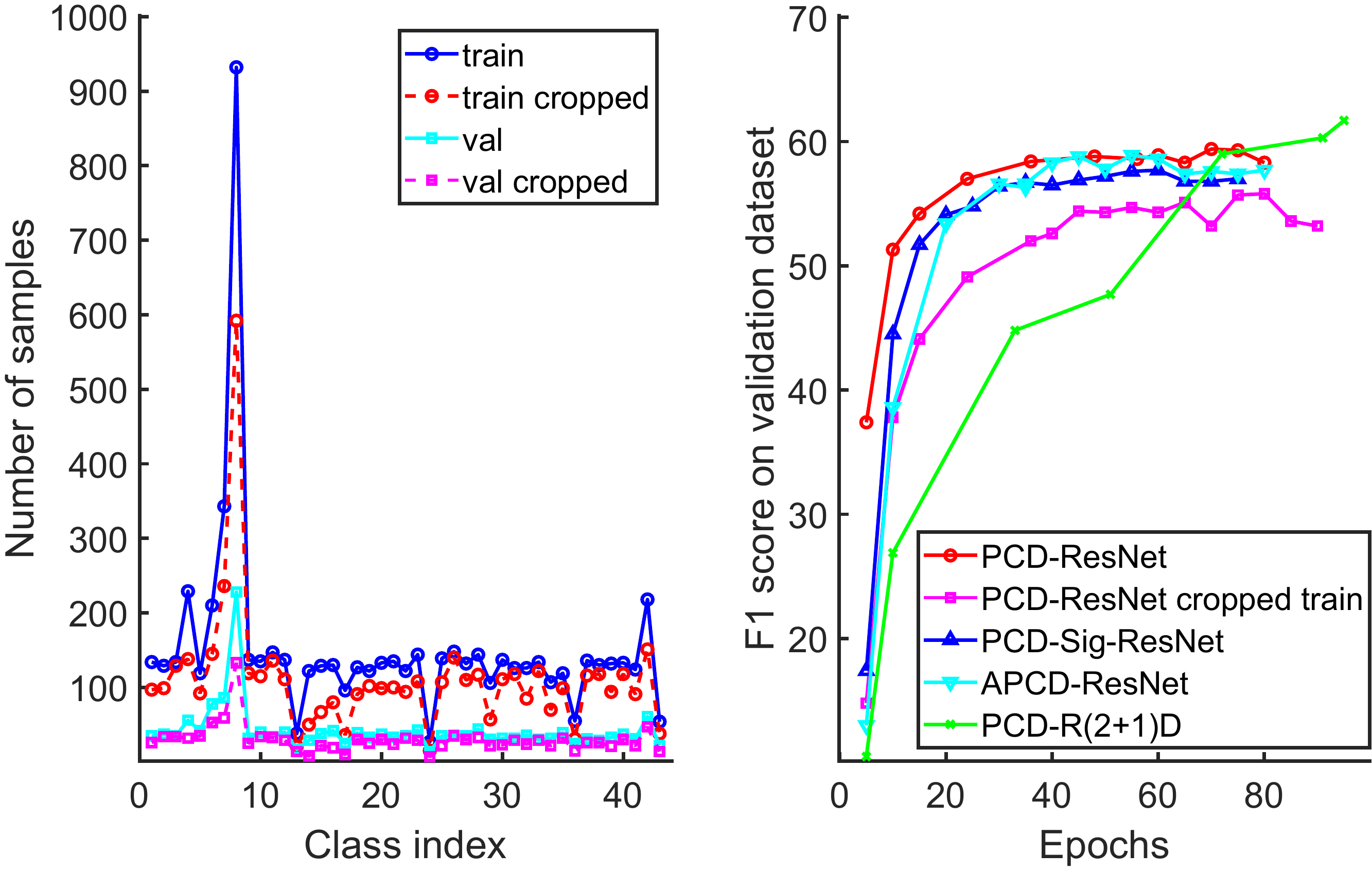}
\end{center}
   \caption{Model performance. A, the distribution of classes in the dataset. Solid blue, training dataset; dashed red, training dataset with random crop applied; solid cyan, validation dataset; dashed megenta, validation data set with random crop applied. B, performance of all models. PCD-ResNet cropped train, PCD-ResNet trained with random crop.}
\label{fig:2}
\end{figure}

\subsection{Alternative architectures}

Each detector in the PCD-ResNet produces a probability distribution in the range of $(0,1)$ by applying a softmax function to two values generated from the fully-connected layer. An alternative is to let the fully-connected layer generate only one value for each class, and use a sigmoid function to regulate the output into the range of $(0,1)$. In this case, we can reduce the number of parameters by $2048\times43+43$. To examine this method, we implemented it as PCD-Sig-ResNet and evaluated its performance on the same validation dataset. We found that this model performed slightly worse than the PCD-ResNet model, but it still works reasonably well considering it has significantly less parameters (Fig. 2B, blue curve).

\section{Attended per-class detection with ResNet (APCD-ResNet)}

\subsection{Architecture}

In the previous section, we showed that while the PCD-Sig-ResNet model has less complexity compared to PCD-ResNet, its performance is also slightly impaired. Therefore, we introduce an attention map \cite{xu2015show} to make the model focus on specific regions and then generate one single predictor for each class based on those regions. In this way, we hope that the reduction in number of parameters will not impair its performance. We hacked into the ResNet and extract the feature map ($1024\times14\times14$) from its layer3, and then obtained an attention map from this feature map with a convolution layer (kernel=3, depth=1, stride=1, padding=1) followed by a sigmoid operation. After that, the extracted feature map is weighted by this attention map and then fed into the rest layers of ResNet-152. We replaced the last fully-connected layer of ResNet-152 with 43 class detectors, each of them is composed with a fully-connected layer ($2048\times1$) followed by a sigmoid function. Then the outputs from all 43 detectors give the final prediction vector ($43\times1$) (Fig. 3).

\begin{figure}[t]
\begin{center}
   \includegraphics[width=0.6\linewidth]{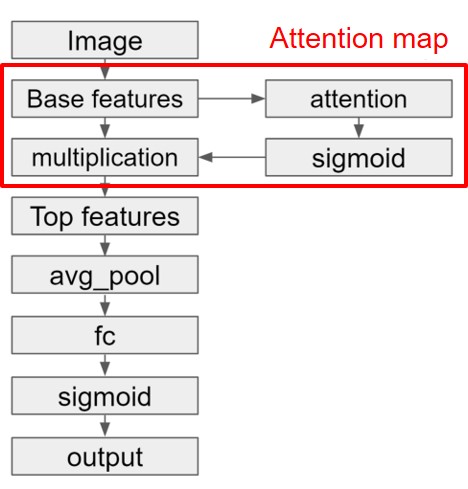}
\end{center}
   \caption{Architecture of APCD-ResNet.}
\label{fig:3}
\end{figure}

\subsection{Training parameters and model performance}

The training parameters were the same as for PCD-ResNet. As shown in Fig. 2B (cyan curve), compared to the PCD-ResNet model, this model converged at a lower but still similar F1 score (58.9 at 55 epochs). Notice that this model has fewer parameters than PCD-ResNet because of the change from softmax to sigmoid.

While both APCD-ResNet and PCD-Sig-ResNet incorporate the same fully-connected and sigmoid layer in their detectors, the attention mechanism makes the APCD-ResNet model outperform the PCD-Sig-ResNet model (Fig. 2B, cyan versus blue curves).

\begin{figure}[t]
\begin{center}
   \includegraphics[width=1\linewidth]{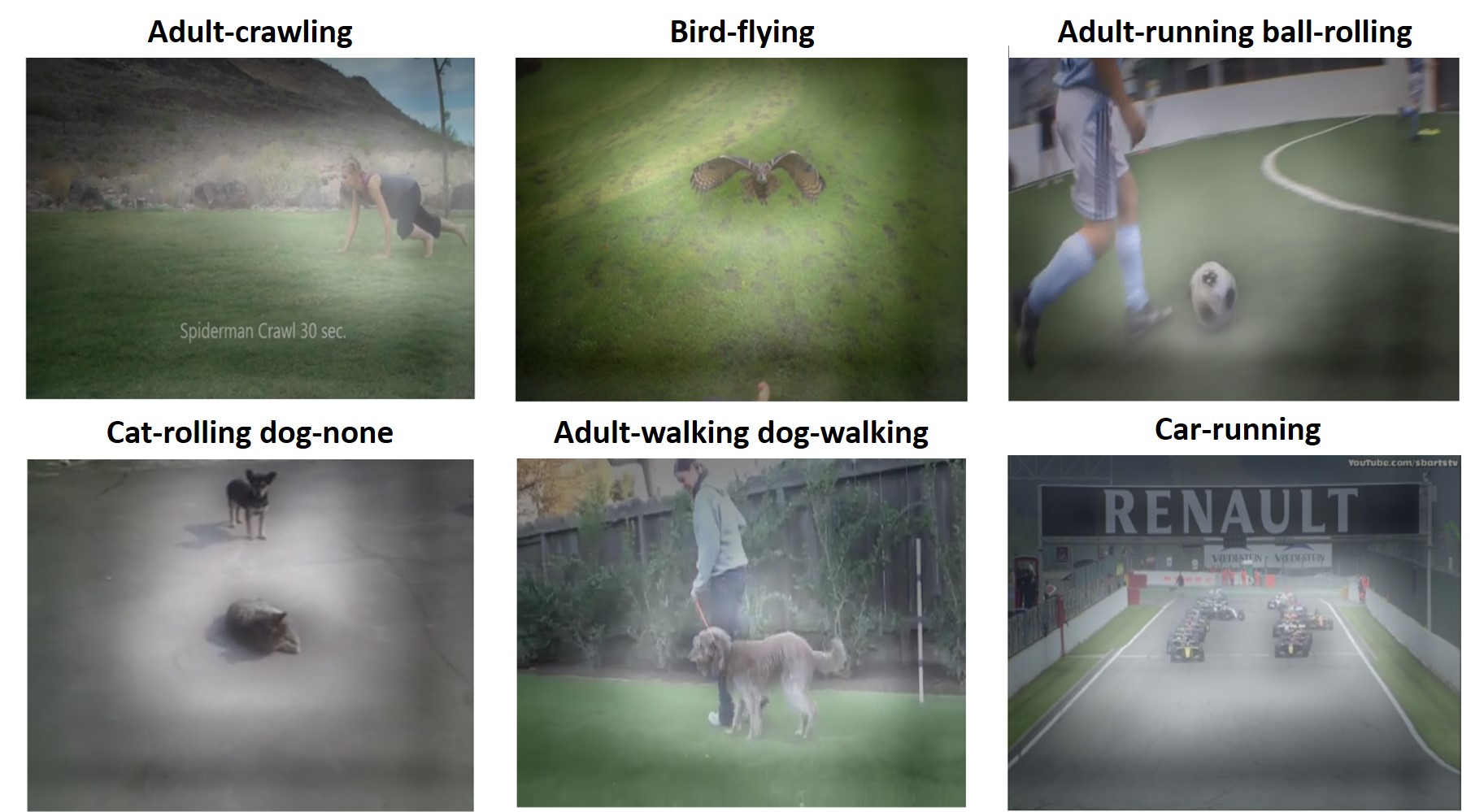}
\end{center}
   \caption{Examples of attention maps generated by APCD-ResNet on the validation dataset. The attention maps allow our network to localize regions of interest while performing actor-action detection.}
\label{fig:4}
\end{figure}

Fig. 4 shows some examples of attention maps generated by APCD-ResNet on the validation dataset. In general, the attention maps are able to capture the regions of interest where the actors and actions locate.

\section{Per-class detection with ResNet-(2+1)D (PCD-R(2+1)D)}

\subsection{Architecture}

The previous two models both focus on single-frame image, yet the nature of actor-action is dynamic, so it would be useful to take advantage of consecutive frames next to the target image. To achieve this goal, we replaced the ResNet-152 backbone in PCD-ResNet with a ResNet(2+1)D model \cite{tran2018closer}, which takes a series of frames (16 frames in total) and performs convolutions in both spatial and temporal domain independently. The design of the 43 independent detectors are the same as those for PCD-ResNet (Fig. 5).

\subsection{Training parameters and model performance}

For the data preparation, we modified A2DDataset such that it returns a given number of consecutive frames with the target image at the center. Other preparation steps were the same as for previous models.

Training parameters including optimization function, learning rate, momentum, weight decay, and loss function were the same as for previous models.

Since the (2+1)D module is much more complicated than previous models, it takes 20 times as long to train this model as it does for previous ones. To speed up the training process, we started with 4 frames for each image, and gradually increased it to 8 frames and finally 16 frames. In addition, we skipped every other frame in the 4-frame case to acquire a long enough time span.

\begin{figure}[t]
\begin{center}
   \includegraphics[width=1.0\linewidth]{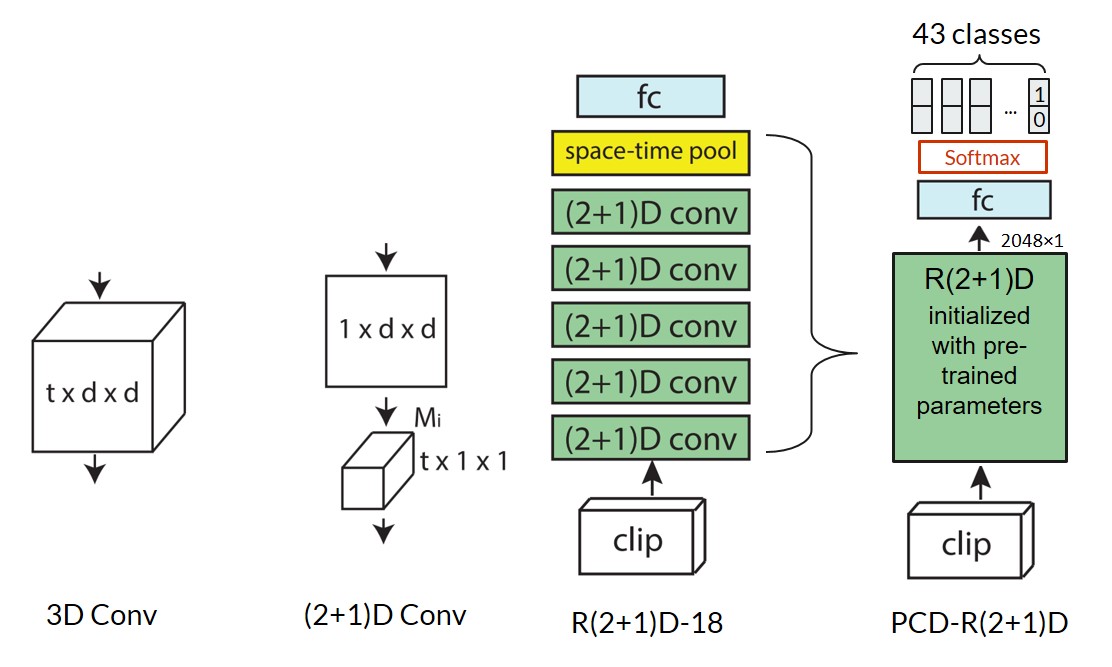}
\end{center}
   \caption{Architecture of PCD-R(2+1)D and its components. A, a traditional 3D convolution layer; B, (2+1)D layer used in the ResNet-(2+1)D model, where convolutions are perform sequentially in pixel space and in temporal space; C, the architecture of ResNet-(2+1)D from \cite{tran2018closer}. D, Architecture of PCD-R(2+1)D.}
\label{fig:5}
\end{figure}

As shown in Fig. 2B (green curve), the PCD-R(2+1)D model evolved much slower than any other models. However, it outperformed the others with a maximum F1 score of 61.7 at 95 epochs. Note that we only trained 95 epochs due to time constraint, but its performance could be higher than 61.7 with more training epochs.

\section{Ensemble: PCD-ResNet + APCD-ResNet + PCD-R(2+1)D}

Since we already built and trained three different models separately, it is possible to obtain higher performance by combining them together \cite{hashem1997optimal,naftaly1997optimal}. Specifically, we first calculated three weights for all models based on their F1 scores, and then obtained a weighted voting score and a weighted average score from their individual predictions, and calculated the mean of the two scores as the final output:
\begin{align*}
    w_k &= \frac{F1_k}{\sum_{i=1,2,3} F1_i}, k\in\{1,2,3\} \\
    v &= \sum_{i=1,2,3} w_i[p_i>0] \\
    m &= \sum_{i=1,2,3} w_i p_i\\
    p &= (v+m)/2 
\end{align*}
where $F1_i$, $w_i$, and $p_i$ are F1 scores, weights, and predictions of the $i$-th model, respectively. $v$ and $m$ each denotes the weighted vote and average scores. $p$ is the final prediction.

\begin{figure}[t]
\begin{center}
   \includegraphics[width=1.0\linewidth]{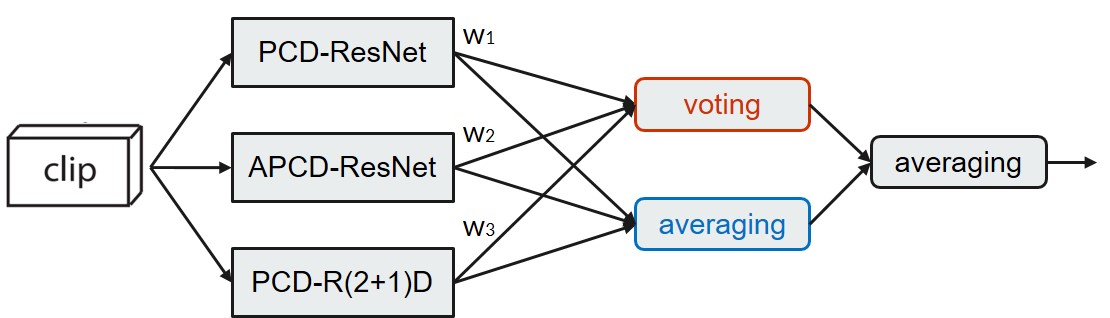}
\end{center}
   \caption{Illustration of model ensemble.}
\label{fig:6}
\end{figure}

The model performance on the validation dataset is shown in Table 1.
\begin{table}
\begin{center}
\begin{tabular}{|c|c|c|c|}
\hline
Model & F1 score & Precision & Recall \\
\hline
PCD-ResNet & 59.4 & 57.9 & 66.3 \\
APCD-ResNet & 58.9 & 57.4 & 65.6 \\
PCD-R(2+1)D & 61.7 & 60.9 & 68.6 \\
Ensemble & 62.9 & 63.2 & 66.8 \\
\hline
\end{tabular}
\end{center}
\caption{Results of three models and the ensemble model.}
\end{table}

\section{Novelty}
\subsection{Converting multi-label classification problem into independent class detection}

For a multi-label classification problem, people usually make use of both ground truth class labels and ground truth bounding boxes or semantic segmentation maps, so they are able to perform single-label classification in single regions of interest (e.g., RCNN \cite{girshick2014rich}, YOLO \cite{yolov3}). Yet in our task, ground-truth bounding boxes or semantic segmentation maps are not available, therefore we cannot directly adopt those models for our task.
On the other hand, typical classification networks are usually built for single-label classification, which output a score vector which sums up to 1 and represents the predicted probability distribution of the target over all possible classes. Obviously, this could not perform well if applied directly to our multi-label classification task.

However, the problem can be easily solved by determining the presence of each class independently. Moreover, instead of training 43 different models for each class independently, which is too costly, we let all 43 detectors share the same feature vector generated by a deep CNN.

\subsection{Assembling three different models}

By assembling all three models proposed in our work, we achieve a higher classification performance. There are two regimes for model ensemble: voting and averaging \cite{hashem1997optimal,naftaly1997optimal}. While the voting method is robust to outliers, it does not take into account the confidence level of each model. The averaging method, on the other hand, considers the difference in confidence level but are easily affected by extreme values. We incorporated both methods in our ensemble model and achieved the best performance among all models proposed in this report.

\subsection{Implementing an attention mechanism}
	
We hacked into a deep CNN and extracted a rich feature map from a hidden layer, then passed the feature map to an attention layer, which consisted of a convolution layer (kernel size of $3\times3$, depth of 1) followed by a sigmoid layer, and finally obtained an attention map. In this way, before the extracted feature map goes through the rest layers of the CNN, it will be filtered by a multiplication with the attention map.

This attention mechanism could potentially extract features from locations most relevant to the actor-action detection, and reduce the complexity of the task.

\subsection{Removing the random crop manipulation }

We showed that random crop reduced the number of occurrences of each class in the training dataset (Fig. 2A). Our experiments show that the crop manipulation impaired the performance on the evaluation dataset.

\subsection{Method to speed up the training of PCD-R(2+1)D}

We made use of the temporal informantion provided in the dataset and improved our model by employing the (2+1)D convolutional module. Because the PCD-R(2+1)D model performs much more computations, it is much slower than our 2D CNNs (a magnitude of about 20). In order to speed up the training, we start from a small number of frames for each image, and gradually increase it when the performance improvement per epoch gets smaller, and we do that repeatedly until reach to 16 frames for each image. Moreover, when the number of frames per image is small, we down sample in time by skipping frames so that the frames we take for each input image still span enough in time. Taken together, manipulating the total number of frames and the sampling rate allows us to speed up the training processing as well as prevent the model from overfitting.

{\small
	\putbib[students/zxu53/egbib]
}
\end{bibunit}

\clearpage

\newpage

\setcounter{section}{0}
\setcounter{figure}{0}
\setcounter{equation}{0}
\setcounter{table}{0}

\title{ A Baseline Comparison for Action Understanding with Multiple Classes of Actors}

\author{Zhongkai Shangguan \hspace{4mm} Yue Zhao \hspace{4mm} Jingwen Wang\\
University of Rochester \\
{\tt\small \{zshangg2, yzhao88, jwang191\}@ur.rochester.edu}
}
\maketitle

\begin{abstract}

ResNet architectures have shown strong performance for multiple-label classification. In this paper we use actor-action dataset (A2D) for action recognition in video dimension. Based on 2D ResNet, we apply the SE (Squeeze-and-Excitation) blocks to the ResNeXt architecture which can efficiently exploit the split-transform-merge strategy and learn feature of different channels. We implement and compare several architectures: Resnet34, PNASNet-5-large, EfficientNet-B7 and SE-ResNeXt101 to this training data set and our results shows that SE-ResNeXt101 have the best performance.

\end{abstract}

\begin{bibunit}[unsrt]
\section{Introduction}
In order to represent simple action combination, most actions recognition task rely on handcrafted features, but those are gradually shown to be efficiently replaced by Convolutional Neural Networks (CNN). 
Many image analysis based pipelines are given, which take raw image as the input and output the classification labels of each image~\cite{krizhevsky2012imagenet}. 
Several CNN architectures have shown state-of-the-arts performance on object recognition \cite{sermanet2013overfeat} \cite{simonyan2014deep}. 
It is comparably challenging when expand the 2D input to a state-of-the-art 3D dimension video data, and one of the most popular method is to use stacked video frames. For actor-action dataset (A2D) \cite{xu2015can}, capturing different actors and their actions consist of segmentation and multi-classification and in this paper we focus on the multi-classification scenario. A2D contains 3782 videos from YouTube, in each of which, objects are annotated with actor-action label. There are overall 43 valid actor-action tuple which is formed by seven actor classes (adult, baby, ball, bird, car, cat, and dog) and eight action classes (climb, crawl, eat, fly, jump, roll, run, and walk) not including the no-action class.

ResNet \cite{he2015deep} (Residual CNNs) is one of the most commonly used architectures or backbones for multi-label tasks, as it can efficiently solve the degradation problem when increasing the accuracy lead to deeper CNN layers. 3D ResNet or spacial ResNet is more commonly used in solving the action recognition problem of the video \cite{kataoka2020megascale} \cite{simonyan2014twostream}. However,3D CNNs heavily rely on the modification of the architecture. In this paper, based on the 2D ResNet architecture, we implemented SE-ResNext101 on A2D and compare with the related architectures, such as EfficientNet-B7.

Our paper introduce the overall implementation with data augmentation, network architecture and optimization methods in Sec. 2, and the experiment of different architectures is shown in Sec. 3. The result and conclusion is shown in Sec.4. Our pre-trained parameters are initialized by ImageNet.

\section{Method}

This section provides details of the training method, including data augmentation, model structure, loss function and optimization method.

\subsection{Data Augmentation}
By doing image augmentation, people can get a better performance when dealing with limited image dataset and avoid over-fit, this will improve the model robustness both on validation and test set.

Some normal augmentation methods like rotation and flip are applied in this project. Before feeding the data to model network, we crop, pad and re-scale the images into (224, 224, 3).

As the images are extracted from videos, motion blur should also be considered as an important part, this project also use Gaussian Blur, Median Blur and Motion Blur for pre-processing images.

Dropout is another powerful technique which aim to make the images more similar to our daily environment. This technique will randomly generate less than five $8\times8$ black squares in order to simulate occlusion of objects in real scenes.

Apart from the above methods, we also modify image contrast by using RandomGamma, HueSaturationValue change, and RandomBrightnessContrast, all these augmentation methods are based on augmentation library \cite{buslaev2020albumentations}.

\subsection{Network Architecture}

We modified and test different CNN architectures, including ResNet34,  PNASNet-5-large, Efficientnet-b7 and SE-ResNeXt101. All models are initialized from pretrained parameters on ImageNet \cite{deng2009imagenet}. Among all this models, SE-ResNeXt101 shows the best performance.

\begin{figure}[t]
\begin{center}
\includegraphics[width=0.8\linewidth]{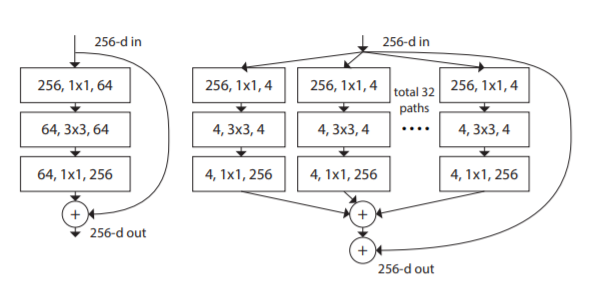}
\end{center}
   \caption{Squeeze-and-Excitation (SE) Block}
\label{fig:resnext101}
\end{figure}

ResNeXt is a simple architecture which adopts VGG/ResNets’ strategy of repeating layers, while exploiting the split-transform-merge strategy in an easy, extensible way. The innovation lies in the proposed aggregate transformations, which replaces the original ResNet's three-layer convolution block with a parallel stack of blocks of the same topology structure (Figure \ref{fig:resnext101}), which improves the accuracy of the model without significantly increasing the magnitude of the parameter.

\begin{figure}[t]
\begin{center}
\includegraphics[width=0.8\linewidth]{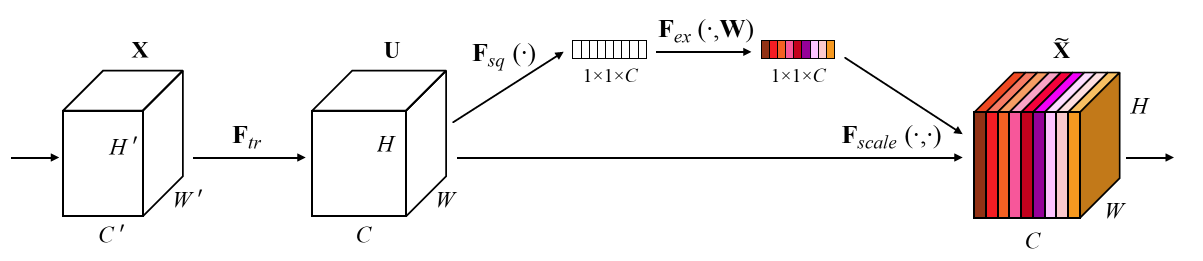}
\end{center}
   \caption{Squeeze-and-Excitation (SE) Block}
\label{fig:seblock}
\end{figure}

Then apply SE (Squeeze-and-Excitation) blocks to ResNext we get SE-ResNext model, the SE block is shown in (Figure \ref{fig:seblock}), the main idea of SE block is by giving different channels a weight which make the model also learn feature of channels. We notice that after convolution, we first apply global maxpooling to squeeze the 2-d channel into a number, then using fully-connected layers learn the weight of each channel, finally multiply the weights to the original 2-d feature.

Finally add a dense layer as a classifier after the Feature extractor. This layer use Mish\cite{mish} as the activation function, and it is followed by a Dropout layer in order to drop redundant features and improve the model robustness.

\subsection{Loss and Optimization}

The loss function is defined as Binary Cross Entropy.
\begin{equation}
    Loss=-w_n [y_n \cdot log\sigma(x_n )+(1-y_n ) \cdot log(1-\sigma(x_n ))]
\end{equation}
The optimization method is SGD (Stochastic Gradient Descent) with two stages, in stage one CosineAnnealingLR is applied with a large initial learning rate, which can help accelerate the converge process. The second stage is used to fun-tune the model in a more meticulous way, with a circle scheduler start with a very small learning rate. Additionally, the training process is initialized with amp(AUTOMATIC MIXED PRECISION), which enable a Mixed Precision training process. It contains two computing type: FP16 (Half-precision 16-bit floating-point) and FP32 (Single-precision 32-bit floating-point).

\section{Experiment}

Two local 2080Ti with 11GB memory were used for the hardware, the Ubuntu 16.04 and 18.04 were used for the system, and the Cuda 10.1 , Python 3.7, PyTorch 1.2.0 were used for the environment. For the performance in the experiment, the three information we collected are precision, recall and F1 score for the pattern recognition. To be specific, precision represents the positive predictive value, which is the fraction of relevant instances among the retrieved instance; recall represents sensitivity, which is the fraction of the total amount of relevant instances that were retrieved and it is the number of correct positive results divided by the number of all relevant samples; F1 score is the measure of a test’s accuracy, which is computed from the precision and recall \cite{experiment}. The Fig.3 shows the training loss and training accuracy of SE-ResNeXt101 with the training hyperparameters Batch size =64, Epoch=200, learning rate =0.05, accumulate=1, Step =10. The left figure shows the training loss of SE-ResNeXt and we can find that the Training loss rapidly decreased to 0.1121 when epoch reached 2, and finally slowly converged to the certain value around 0.03-0.04. The right figure shows the training accuracy, and it has huge accuracy improvement from the epoch 1 to epoch 28, and then coverage slowly to the epoch 192 eventually. The sinusoidal shape has been observed for the training accuracy since the optimization method SGD with two stages were used and they are CosineAnnealingLR and MutiStepLR. The training Time for SE-ResNeXt101 is average 90s per epoch and the training loss is 0.1258 when it finishes the first epoch. The best performance at epoch 192 and the number of iterations is 3300.

We also did the comparison with different models such as Resnet34, PNANet 5 large, Efficient Net-B7 (with and without augmentation), SE-ResNeXt101(with and without augmentation), and the result in fig.4 shows that SE-ResNeXt101 with the activation function Mish has the best performance.

\begin{figure}[t]
\begin{center}
\includegraphics[width=0.8\linewidth]{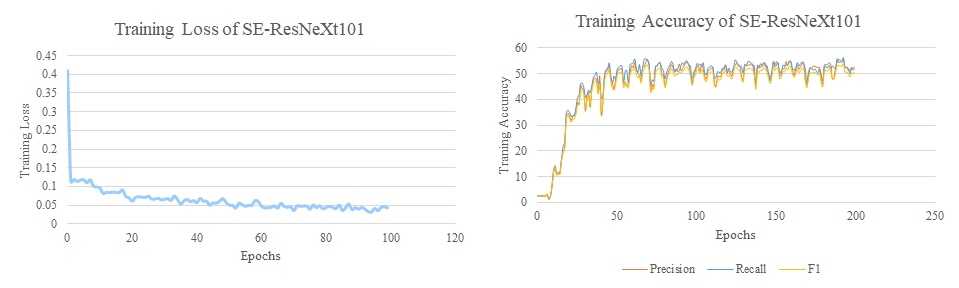}
\end{center}
   \caption{Training Loss and training accuracy of SE-ResNeXt101}
\label{fig:Figure3}
\end{figure}

\begin{figure}[t]
\begin{center}
\includegraphics[width=1.0\linewidth]{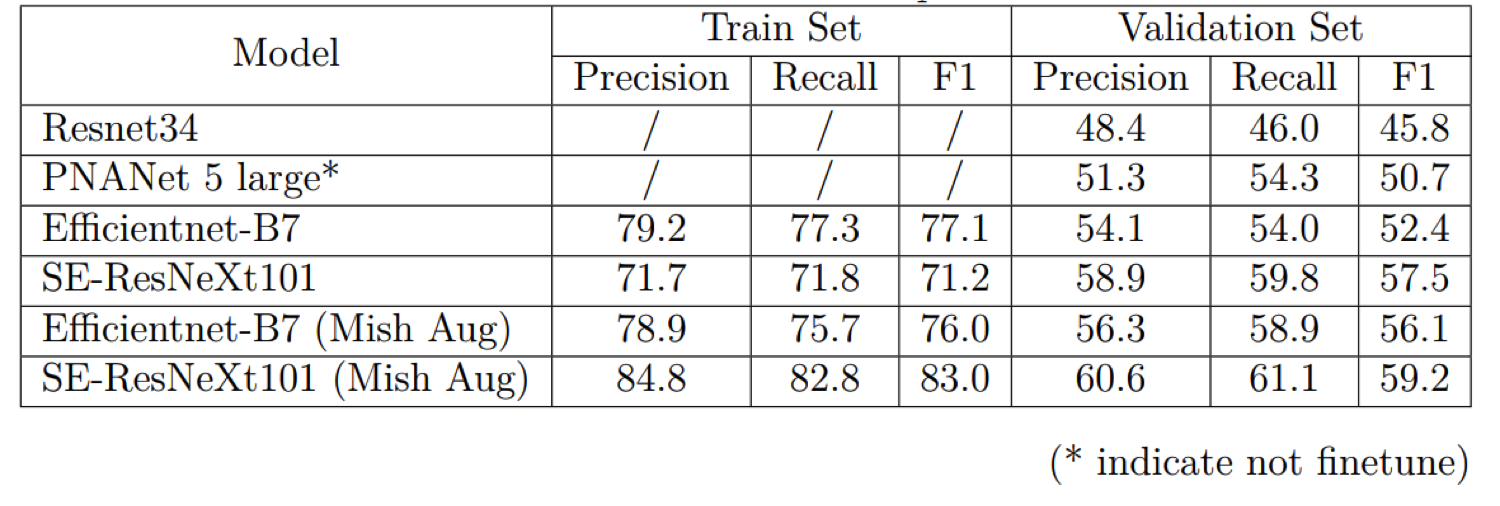}
\end{center}
   \caption{The results comparison with different models}
\label{fig:Figure4}
\end{figure}

\section{Conclusion}

SE-ResNeXt101 shows better performance for the validation set with the precision of 60.6, recall of 61.1 and F1 score of 59.2. Our experiments shows that the optimized model architecture contributes to the accuracy of the results. Proper data augmentation can also improve the performance,while other image pre-processing methods such as vertical flip, transpose and all sharpen methods (sharpen, CLAHE) are not useful. The smoothing training process methods, including optimizer selection, two stage for training using different scheduler and accumulator, also decrease the training time and contribute to higher accuracy. For future works, we can try new models and activation method as well as blend models. Also, combining two learning rate into one can be a good direction for new approaches to this task.

{\small
	\putbib[students/zshangg2/egbib]
}
\end{bibunit}
\clearpage

\setcounter{section}{0}
\setcounter{figure}{0}
\setcounter{equation}{0}
\setcounter{table}{0}

\title{Multi-label Action Actor Classification}
\author{
Rohan Sharma \hspace{4mm} Surya Iyer \hspace{4mm} Ajinkya Deshmukh \\
University of Rochester\\
{\tt\small \{rshar27,siyer9,adeshmu2\}@ur.rochester.edu}
}
\maketitle

\begin{bibunit}[unsrt]
\begin{abstract}
    Actor and action by mapping videos with the descriptors remains an actively researched area in the field of machine vision, with much progress attributed to the novel use of techniques such as attention, bounding box identification and labelling, semantic segmentation and a combination of novel architectures that leverage the advancements in techniques and processing capacities. In this project, we explored and experimented with the possibility of action-actor prediction without leveraging bounding box labelling and semantic segmentation, using some novel approaches of our own. We conduct our experiments on the A2D \cite{A2D} dataset containing 3782 video clips from YouTube annotated with 43 actor-action pairs.
\end{abstract}

\section{Motivation and Related Studies}

\begin{figure}
    \renewcommand{\figurename}{Figure}
    \includegraphics[width=\linewidth]{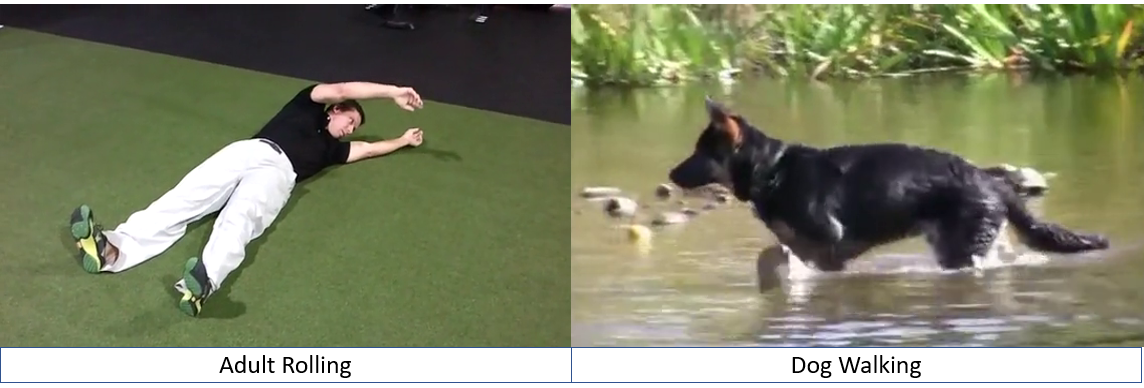}  
    \caption{Sample Images and Tags}
    \label{fig:sample}
    \end{figure}

    The goal of this study is to identify actor-action pairs from frames of a video without explicit usage of spatio-temporal localization such as bounding boxes. The A2D dataset contains 43 labels of actor-action, which are valid for the set of video clips. As an example, the figure \ref{fig:sample} is illustrative of the pairs "adult-rolling" and "dog-walking". Past studies have focused on joint inference over the labels and have indicated the advantages of a joint inference over marginal inferences of actors and actions (\cite{kalogeiton2017joint} and \cite{xu2016actor}). \cite{cheema2012doing} have used bilinear models for two factor classification using Singular Value Decomposition. \cite{kalogeiton2017joint} uses a two stream model and a fusion of the streams that use the RGB input from videos and optical flow, using separate architectures for the two flows. We have implemented as one of our approaches, the use of optical flow lines for better classification of actions, using both videos and images as the inputs. 
    
    Xu et al. \cite{xu2016actor} on the other hand have proposed a Grouping Process Model (GPM) which uses the video content to model a dynamic and continuous process of information exchange of a labeling Conditional Random Field and a supervoxel hierarchy. They have recognised the contrast between actors and actions in that, actor classification relies merely on the spatial interactions between pixels, whereas actions are inferred using spatio-temporal information. We have, in addition to using spatio-temporal information using videos, attempted to infer actions by incorporating temporal information from images into the spatial space using optical flow lines. Additionally, we have also used temporal information from videos, and presented a comparative study of the two approaches. Overall, we have tried an alternative approach towards solving the problem of action-actor joint inference and experimented with marginal inferences to further improve the joint inferences.

\section{Models}

\subsection{Actor Action Baseline Prediction Model (Architecture 1)}

\begin{figure}
    \renewcommand{\figurename}{Figure}
    \includegraphics[width=\linewidth]{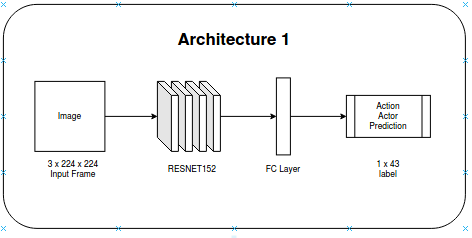}  
    \caption{Baseline Model}
    \label{fig:arch1}
    \end{figure}

   In this subsection we introduce the baseline prediction model for the joint classification of the actor-action pair. We use a pretrained RESNET-152 model and train it end-to-end to predict both the actor and action classes jointly. The model is illustrated in the figure \ref{fig:arch1}. Once the model converged in the evaluation metrics, we use a Learning Rate Scheduler with Cosine Annealing, and to fine-tune the final layers, to boost the performance and saw marginal improvements. Much of our approach has been to improve upon this performance, using new methods for variable generation and combinations of architectures.
   
\subsection{Actor Prediction Model}

    We use the pretrained RESNET-152 architecture to predict the actor class independently. This model achieved a significantly high F1 score. We hypothesise the reason for this remarkably high performance as the specialization of the RESNET-152 in object identification. The results are illustrated in table \ref{Table:performance ml}.

\subsection{Action Prediction Models}

    Action prediction has been an actively researched domain. In this sub-section we try to develop a strong action predictor model for use in combination with the actor predictor described above. We hypothesise that the use of a good action predictor model in conjunction with the baseline actor predictor would help us improve upon the overall performance of the task.

\subsubsection{Variant 1 (Baseline Action Predictor)}
    This variant uses a pretrained RESNET-152 architecture to predict only the actions from images and videos. Our motivation here was to set a baseline standard to compare against. In our experiments with the actor-action label classification, we found that the baseline scores for action only prediction suffered due to a lack of informative variables. Subsequent models in this sub-section attempt to address this lapse through the use of various techniques to generate meaningful variables for the action only classification, with the goal of improving upon the overall score for the actor-action classification task.

\subsubsection{Variant 2}
    
\begin{figure}
    \renewcommand{\figurename}{Figure}
    \includegraphics[width=\linewidth]{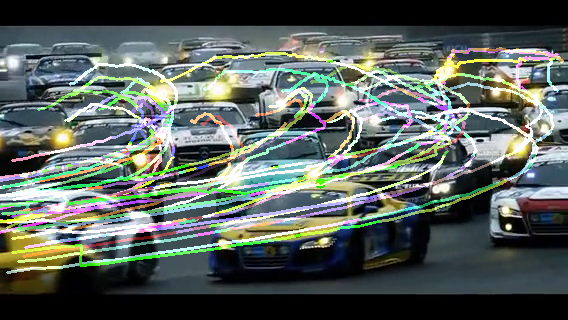}  
    \caption{Optical FLow Lines}
    \label{fig:Optical Flow}
    \end{figure}

    In this variant for action predictor, as a pre-processing step, we use the Shi Tomasi corner detection \cite{shi1994good} and Lucas Kanade Optical Flow detection \cite{lucas1981iterative} to generate flow lines on images. An illustration of the input is shown in figure \ref{fig:Optical Flow}, which indicates using flow-lines, the motion of the cars in a "car-running" image.

    As a sub variant, we also used the flow lines without the raw images to predict actions. These input images were then used to train a RESNET-152 architecture. We found that the flow lines don’t lead to improvements over the existing action predictor from Variant 1. The results from predicting actions from these images are illustrated in the Table \ref{tab:Performance}. We hypothesise that the lapse in the performance may be due to additional noise added to raw images and due to miscellaneous flow lines formed due to camera motion. The performance of this model may be improved upon by controlling for camera motion, or by using still camera videos.
    
\subsubsection{Variant 3}

\begin{figure}
    \renewcommand{\figurename}{Figure}
    \includegraphics[width=\linewidth]{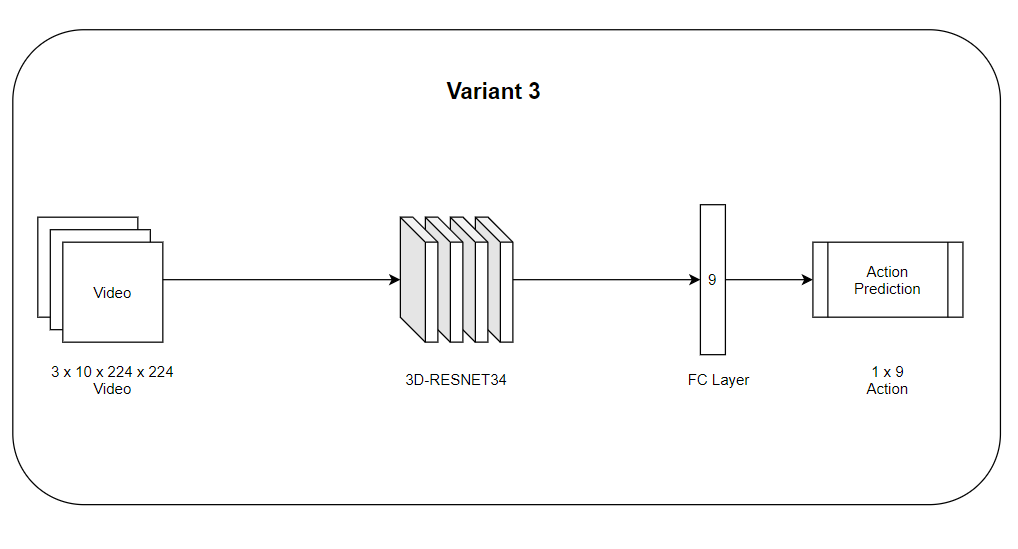}  
    \caption{Action prediction using  3D Resnet (Variant 3)}
    \label{fig:Variant 3}
    \end{figure}

    This variant for action classification uses temporal information from the videos in the form of image frames. For each image we process 5 adjacent frames before, and after the current image, to incorporate temporal information about the motion of objects. This set of frames was used as input for the 3D RESNET-34 \cite{hara3dcnns} to predict the action class. We hypothesised that the use of temporal variance in the positions of objects, in conjunction with a 3D convolution architecture would help improve upon the baseline scores for action prediction. However, the F-1 score achieved by the model was comparatively lower. The architecture for this variant is illustrated in the figure \ref{fig:Variant 3}.

\subsection{Actor Action Prediction Models}
 
    In this sub-section we describe the various attempts to improve upon the baseline for the actor-action model. The attempts utilise the action detectors described above in conjunction with the actor model. 

\subsubsection{Architecture 2}
\begin{figure}
    \renewcommand{\figurename}{Figure}
    \includegraphics[width=1\linewidth]{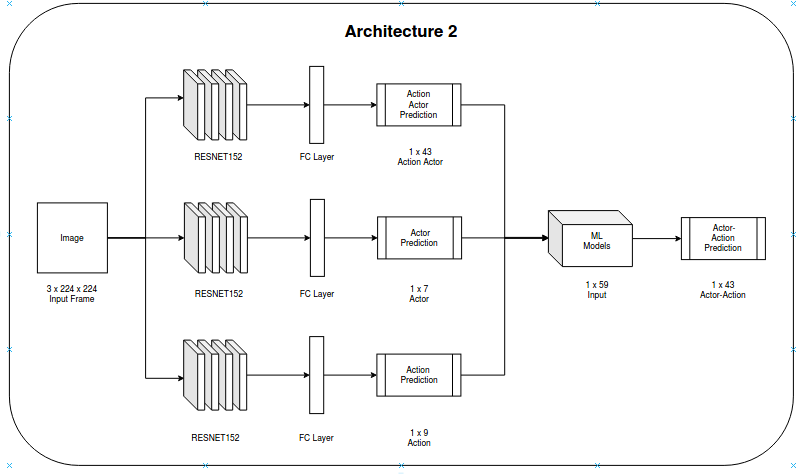}  
    \caption{Statistical ML approach using individual predictions}
    \label{fig:arch2}
    \end{figure}

    As the first variant to this architecture, we use the prediction vectors from the baseline Actor and Action prediction models, and architecture 1 for the joint prediction and subsequently use several statistical machine learning multi-class classifiers, on the output vectors of these models, to predict the main variable of actor-action. This model is illustrated in the figure \ref{fig:arch2}. The ML classifiers we used included Decision Trees, Naive Bayes, Linear Discriminant Analysis, Quadratic Discriminant Analysis, Support Vector Machines and Logistic Regression, in addition to others. As a pre-processing step for the data, we standard-normalized the input to the ML classifiers and tried one-vs-all classification, along with ensemble based methods such as Boosting (for Decision Trees). 
  
    \begin{figure*}
    \centering
    \renewcommand{\figurename}{Figure}
    \includegraphics[width=0.8\linewidth]{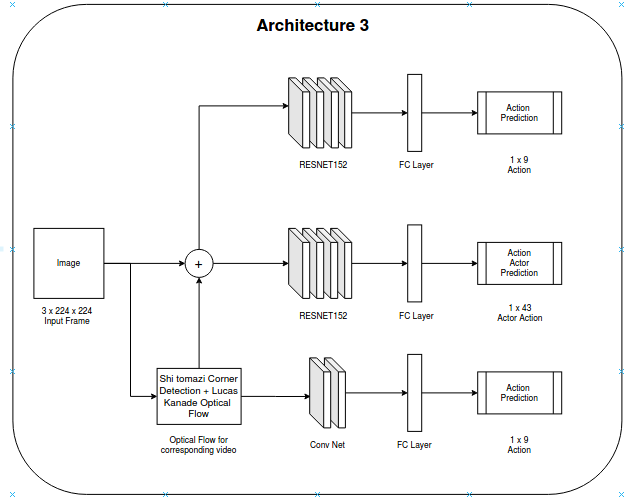}  
    \caption{Optical flow to predict Actor-action pair}
    \label{fig:Arch3}
\end{figure*}

    The performance of this model was fair, in comparison with the baseline, with marginally lower scores. The best result achieved was through Linear Discriminant Analysis as is illustrated in the table \ref{Table:performance ml}.
  
    As the second variant to the model, we tried using high-dimensional FC layers at the three models of dimensions $128\times1$ each, and subsequently projecting them onto a hidden layer of dimension $128\times1$, and finally applying a FC layer to make predictions. In this variant, the results were comparable to the first variant, as is illustrated in the table \ref{tab:Performance}.

    We observe that the performance is hindered due to the low performance of the action prediction model, and we expect the performance of this model to improve with the performance of the action prediction model.

\subsubsection{Architecture 3}

\begin{figure*}
    \renewcommand{\figurename}{Table}
    \includegraphics[width=\textwidth]{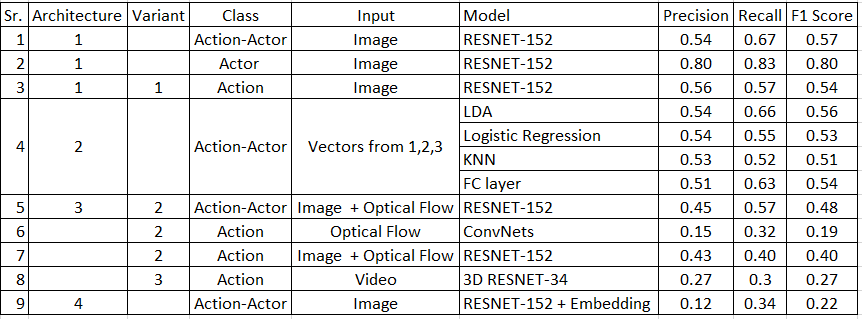}  
    \caption{Performance of Architectures}
    \label{tab:Performance}
\end{figure*}

    In this Architecture, we used the input to variant 2 of the action prediction model (images with optical flow lines) to predict the joint actor-action class using the pretrained RESNET-152. We wanted to compare this model to the baseline model of predicting actions using raw images. We found that the flow lines don’t lead to improvements over the existing action predictor from Model 1. The results from predicting actor-action classes from these images are illustrated in table \ref{tab:Performance}. The model is illustrated in the figure \ref{fig:Arch3}

\subsubsection{Architecture 4}

    In this architecture, we tried predicting the actor-action pair using additional information from pre-trained glove 6B 50D word-embeddings. In this model, we used the actor predictor and action predictor models from Architecture 1, and replaced the FC layers with new layers of dimensions $1\times50$ each. Subsequently, we concatenated these layers and added the concatenated mean values of embeddings from actions and actors to the corresponding FC layers. Later, we used the newly constructed FC layer and subsequent FC layers, with Relu non-linearity to predict the main actor-action variables.
    
    \begin{figure}[t]
    \renewcommand{\figurename}{Figure}
    \includegraphics[width=\linewidth]{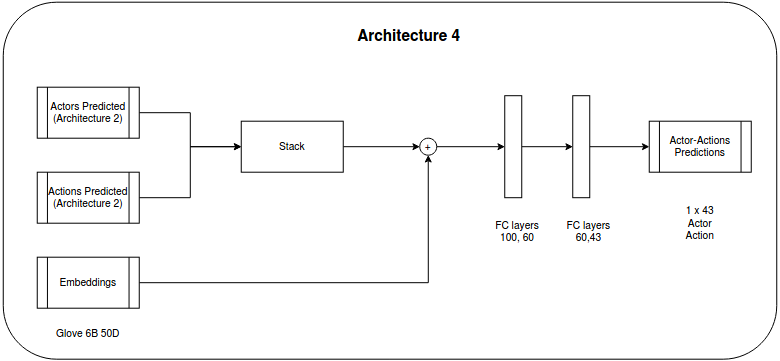}  
    \caption{Actor-Action Prediction (Pretrained Embeddings)}
    \label{fig:arch4}
    \end{figure}

    The idea here was that the labels have relevant semantic information that can be exploited using word embeddings. This method was used to exploit the information from the semantic similarity of words in the labels in high dimensional spaces, to make better predictions for frequently co-occurring words in these labels. As an example: ”dog-running” is a more likely label than ”dog-flying” due to the high co-occurrence of the words "dog" and "running". This co-dependency of words in labels can be exploited using glove embeddings, which are pre-trained using co-occurrence statistics from a large corpus of text.
    
\begin{figure}[t]
    \centering
    \includegraphics[height = 5cm]{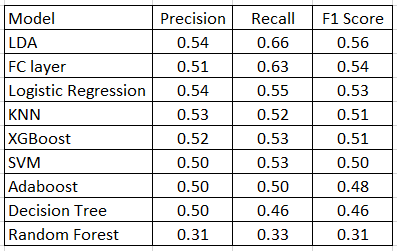}
    \caption{Performance of ML models on Architecture 2}
    \label{Table:performance ml}
\end{figure}

\section{Performance}

    We observe through our experiments with various architectures and techniques, that the overall performance of the Architectures 1 and 2 were the best and close to each other. We also observe that use of additional temporal information to improve upon the actions classification does not improve upon the naive classifier and requires further analysis in terms of the quality of variables used and methods of analysis. As an improvement over the performances, we suggest the use of weight standardization of the convolution layers \cite{qiao2019weight}, to improve the deep learning capabilities of the architectures illustrated. Also, use of a RNN based architecture using an encoder-decoder framework, which may incorporate pre-trained word embeddings may generate coherent mappings to the classes for this task.

{\small
	\putbib[students/rshar27/egbib]
}
\end{bibunit}

\clearpage

\newpage

\emptythanks

\setcounter{section}{0}
\setcounter{figure}{0}
\setcounter{equation}{0}
\setcounter{table}{0}

\title{Evaluation of Transfer Learning on Actor-Action Classification}

\author{Raunak Mahalik \hspace{4mm} Srishti Singh \hspace{4mm} Jayant G Rohra\\
University of Rochester\\
{\tt\small \{rmahalik, ssingh59, jgiridha\}@ur.rochester.edu}
}

\maketitle

\begin{abstract}
   We propose a method for the classification of A2D data-set using transfer learning and deep neural networks. The A2D data-set contains a total of 3728 videos from YouTube, from which frames are extracted resulting in 4750 training images, 1209 validation images and 1044 test images having 43 actor-action paired labels. Each video on average has 120 frames and out of these we have annotations for 3-5 frames per video, thus constituting a sparsely labeled data-set. In this report we describe the technical details of all the experiments performed to achieve a model that performs the task of actor-action classification effectively. Experimental results demonstrate that the approach of transfer learning with freezing a few layers and unfreezing other layers in the pre-trained network in combination with fully connected layers and batch normalization can significantly improve the actor-action classification.
\end{abstract}

\begin{bibunit}[unsrt]
\section{Introduction}

Actor-Action Classification is a multi-class multi-label classification problem where all the actors and actions in a video are classified using a set of 43 actor-action labels like baby-crawling, bird-flying, etc. The A2D data-set contains sparsely labeled data, hence using a pre-trained model and utilizing its weights in conjunction with a linear layers to further train the model significantly improve the performance. A model that has already been trained using another data-set, is referred to as a pre-trained model. Usually, the learned features can be transferred to models which solve similar problems as the original model. This concept of using this trained model for another data-set is known as transfer learning.

Computer vision problems usually suffer from a lack of labelled training data. The process of collecting thousands of training images and labelling them is usually time consuming and costly. Even if a training corpus of accurately labelled data is available, the time consumed while training the network is a bottleneck. Hence, processes like semi-supervised learning, transfer learning, etc. are quite popular when dealing with computer vision problems.

\section{Related Work}

Quite some work has been done in the multi-label image classification space. Recent work \cite{liu2018multi} proposed deep learning frameworks for multi-label classification that can efficiently handle weakly-supervised detection tasks which do not even have any object annotations or bounding boxes. In fact, experiments performed with the A2D data-set \cite{Xu_2015_CVPR} demonstrated that classification made over a joint combination of actors and actions outperform the classifications made on each of these categories individually. However, the problem involves dealing with sparsely labelled data which has been an age-old problem to computer vision researchers.

The most commonly used technique to mitigate the lack of labeled data in a computer vision problem is to leverage the visual features learnt by pre-trained networks. These pre-trained networks learn from an already labelled, large and generic data-set, thus performing reasonably well in practice. The ResNet152 \cite{he2016deep} is one such state-of-the-art architecture that has consistently yielded good results in many visual recognition tasks. However, directly leveraging pre-trained techniques on a new task may lead to the model falling short of the ability to understand different contexts and quickly grasp new concepts. Hence, we utilize transfer learning mechanisms by freezing the weights at a few initial layers and training the rest. Liu et al. \cite{liu2019novel} studied the complex problem of scene classification and carried out an extensive study which involved the utilization of data augmentation techniques in fusion with transfer learning on the ResNet, thereby proving that freezing initial layers and fusing them with a few more for classification by softmax regression performs significantly better than the state-of-the-art pretrained models. In recent times, many other transfer learning mechanisms have been proposed. One such being, SpotTune \cite{Guo_2019_CVPR} wherein a policy network is used to make decisions on whether to pass the image through fine-tuned layers or pre-trained ones.  However, these methods prove to be computationally costly and significantly increase the complexity of the network. 

\begin{figure*}
\begin{center}
\fbox{\includegraphics[scale=0.5]{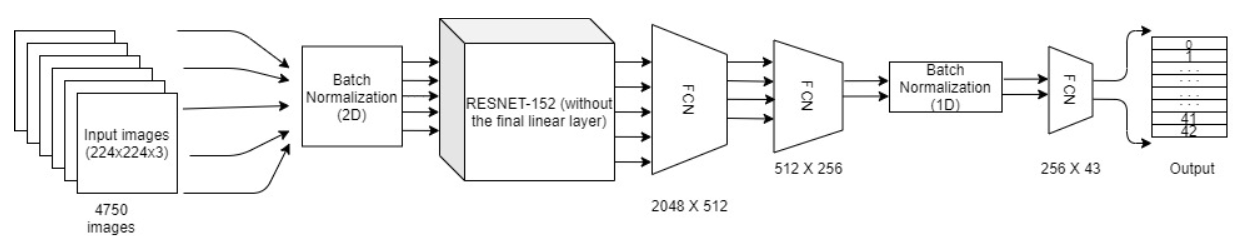}}
\end{center}
   \caption{Network Architecture.}
\label{fig:short}
\end{figure*}

\section{Network Architecture}

The network consists of a pre-trained Resnet-152 network in ad-junction with batch normalization and fully connected layers. It is worthwhile mentioning that deep learning networks progressively learn specialized features in higher layers. For instance, a model would detect edges in the initial layers, shapes in the middle layers and high-level features in the final layers. Usually, the features learned in the final layers are specific to the training data for a specific task. Hence using the weights from these higher layers while training the network on a new set of completely different data can be detrimental. To avoid this, the final layer of the Resnet-152 was chopped off and a new feed forward linear layer was introduced.

Moreover, the weights of the Resnet-152 layers were frozen up to layer5 and the weights of rest of the layers were continuously updated. The input images were 2D batch normalized before feeding into the Resnet-152 network. The output features from the modified Resnet-152, 1000 features, were then passed on to a series of fully connected layers with decreasing number of nodes as moved forward, producing an output of 256 features. These features are then batch normalized along a single dimension and passed on to the final fully connected layer which maps the output to 43 actor-action paired labels. All the actors-actions sets were given a probability in this layer and those crossing a certain threshold were given a value of one, while others were set to zero in a list of length 43 corresponding to all possible actor-action pairs. The Binary Cross Entropy Logits loss was used in combination with the Stochastic Gradient Descent optimizer having a constant learning rate. The network architecture is shown in figure 1. The network was evaluated using three evaluation metrics, precision, recall and F1 score.

\section{Experiments}

The model proposed in the previous section was achieved after performing multiple experiments with different architectures. In this section, we will discuss the results achieved and challenges faced with different models. First, the model was trained using a simple deep convolutional neural network with four layers, followed by two FCN layers. Since, the data was sparsely labeled, this network had poor performance. Hence, transfer learning was used for further training.

Different network models were built using various pre-trained networks including VGG-16, Resnet-152, Resnet-50, Resnet-34 and Alexnet. Among these pre-trained models, Resnet-152 was found to be performing the best. Thus, Resnet-152 pre-trained model was used for further experimentation. The final layers of the pre-trained models are specific to the features in the data-set it was trained on, therefore the last layer of the pre-trained was removed resulting in the output of 1000 features from the pre-trained model. A sequence of fully connected layers were then placed to map these features to the output. The number of layers, loss method, optimizer algorithm, batch size, number of epochs and learning rate were tuned by evaluating the precision, recall and F1 score.

Auto-encoders are usually used for data de-noising and dimensionality reduction, thus auto-encoders were added post pre-trained network, before mapping the output to labels. But there wasn't any significant improvement in the model and auto-encoders add additional computation therefore, they were removed from the network.

Freezing the initial layers of pre-trained network reduces the computation time by avoiding back-propagation through all the layers and mitigates model over-fitting. Thus, the first few layers of the network were frozen and found the best performance with initial five frozen layers. The other layers in the network were trained for all the epochs.

The model hyper-parameters like learning rate, number of epochs, batch size, number of FCNs, optimizer, etc. were tuned to identify the model highest evaluation metrics values.The Adam and SGD optimizers as well as BCELoss and BCEWithLogitsLoss were used in combination with different number of epochs, batch size and learning rates to conclude that the SGD with BCEWithLOgitsLoss outperforms all other models. Though Adam optimizer auto tunes the learning rate, it was observed that the SGD optimizer performs better than tha Adam for this classification problem. The evaluation metrics achieved with each of these models is shown in figure 2.

\begin{figure}
\begin{center}
\includegraphics[scale = 0.68]{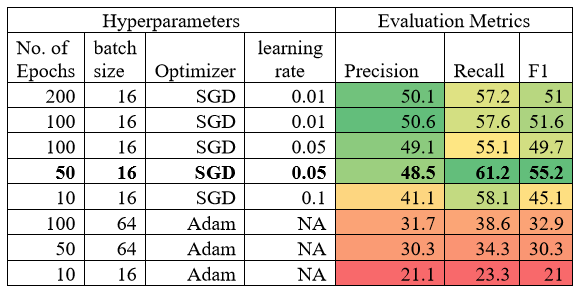}
\end{center}
   \caption{Evaluation metrics for different hyper-parameters.}
\label{fig:eval metrics}
\end{figure}

\section{Conclusion and Results}

It can be concluded that the transfer learning gives better performance in case of sparsely labeled data-sets. The best results were achieved by using the Resnet-152 pre-trained network for the actor-action classification. Figure 3 shows the performance of different models as hyper-parameters varied. The performance metrics achieved for the validation set of the A2D dataset have the precision equal to 48.5\%, recall equal to 61.2\% and the F1 score was 55.2\%.

The evaluation metrics for the test set decreased by some amount as compared to validation set with precision of 46.72\%, recall of 59.47\% and the F1 score achieved was 49.63\%. This was achieved by using BCEWithLogitsLoss along with SGD optimizer having a learning rate of 0.05 with a batch size of 16 at 50 epochs after which model starts to over-fit on training data.

\begin{figure}[h]
\begin{center}
\includegraphics[scale = 0.52]{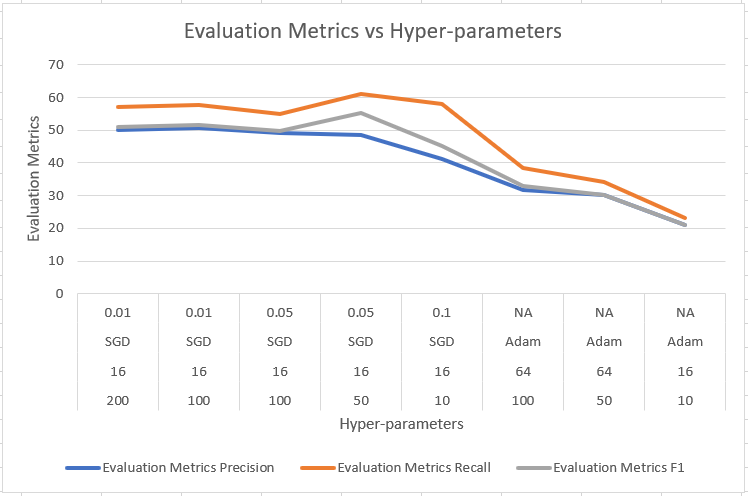}
\end{center}
   \caption{Evaluation metrics against hyper-parameters.}
\label{fig:evaluation metrics graph}
\end{figure}

{\small
	\putbib[students/ssingh59/References]
}
\end{bibunit}
\clearpage

\clearpage
\newpage

\setcounter{section}{0}
\setcounter{figure}{0}
\setcounter{equation}{0}
\setcounter{table}{0}
\title{Evaluating Object Proposal-based and Multi-Scale Fusion-based Methods for Actor-Action Classification on Still Images}

\author{Yipeng Zhang \hspace{4mm} Tongyu Yang \hspace{4mm} Xuan Wen\\
University of Rochester\\
{\tt\small \{yzh232, tyang20, xwen5\}@u.rochester.edu}
}

\maketitle
\begin{abstract}

In this report, we compare the performance of two proposed methods on the multi-label multi-class actor-action image classification task on the A2D dataset. Our first approach utilizes object bounding box proposals to perform object-level classification; our second method performs fusion of predictions on feature maps of different scales with the help of a self-attention module. Experimental results demonstrate our multi-scale fusion-based method significantly outperforms the object proposal-based method and the baseline model.
   
\end{abstract}
\begin{bibunit}[unsrt]
\section{Introduction}
Multi-label multi-class image classification task requires the model to determine the class predictions of an image, where each image might have more than one label. The A2D dataset is an actor-action video classification dataset proposed in~\cite{xu2015can}. In this report, we only use the still image classification part of the dataset. It contains 43 valid actor-action labels such as ``cat climbing'' and ``adult crawling''. It has 4750 training images, 1209 validation images, and 1044 test images, where only the training and validation sets are released as a challenge. 

There are two major challenges of this task. First,  actions are performed along some time period; such information is hard to retrieve from a single image. Previous studies have shown the capability of deep neural architectures on actor-action semantic segmentation on videos~\cite{xu2016actor, Kalogeiton_2017_ICCV, yan2017weakly}, which has consecutive frames that add information to the action performed, whereas in this task, only one single image is given as input. Second, images may contain more than 1 actor, which are of different sizes. The model needs to perform some object detection by nature, although the task is image classification. 

In this report, we propose two methods that mainly focus on overcoming the second challenge. The first method generates object proposals to directly perform object detection before passing the detected objects to the classification network. The second method takes feature maps of different scales and uses a  self-attention module to locate the objects. We also compare our methods with a baseline model, the popular ResNet architecture~\cite{he2016deep}. 

\section{Method}

\subsection{Baseline}

There are many successful image feature extraction deep learning models proposed in the last decade~\cite{simonyan2014very, he2016deep, szegedy2016rethinking, krizhevsky2012imagenet}. One of the most effective and widely used architectures is the deep residual network (ResNet) proposed in~\cite{he2016deep}, where residual connections are added within different parts of the network in order to maintain the information flow. We adopt the ResNet152 model as our baseline method. We add another fully connected layer at the end of the network to turn the length of the original vector output from 1000 to 43, which is the number of actor-action classes in the A2D dataset. We apply element-wise sigmoid activation function to output the per-class confidence score vector, enabling the model to do multi-label multi-class classification.

\subsection{Object Proposal}
\begin{figure}
\begin{center}
\includegraphics[width=\linewidth]{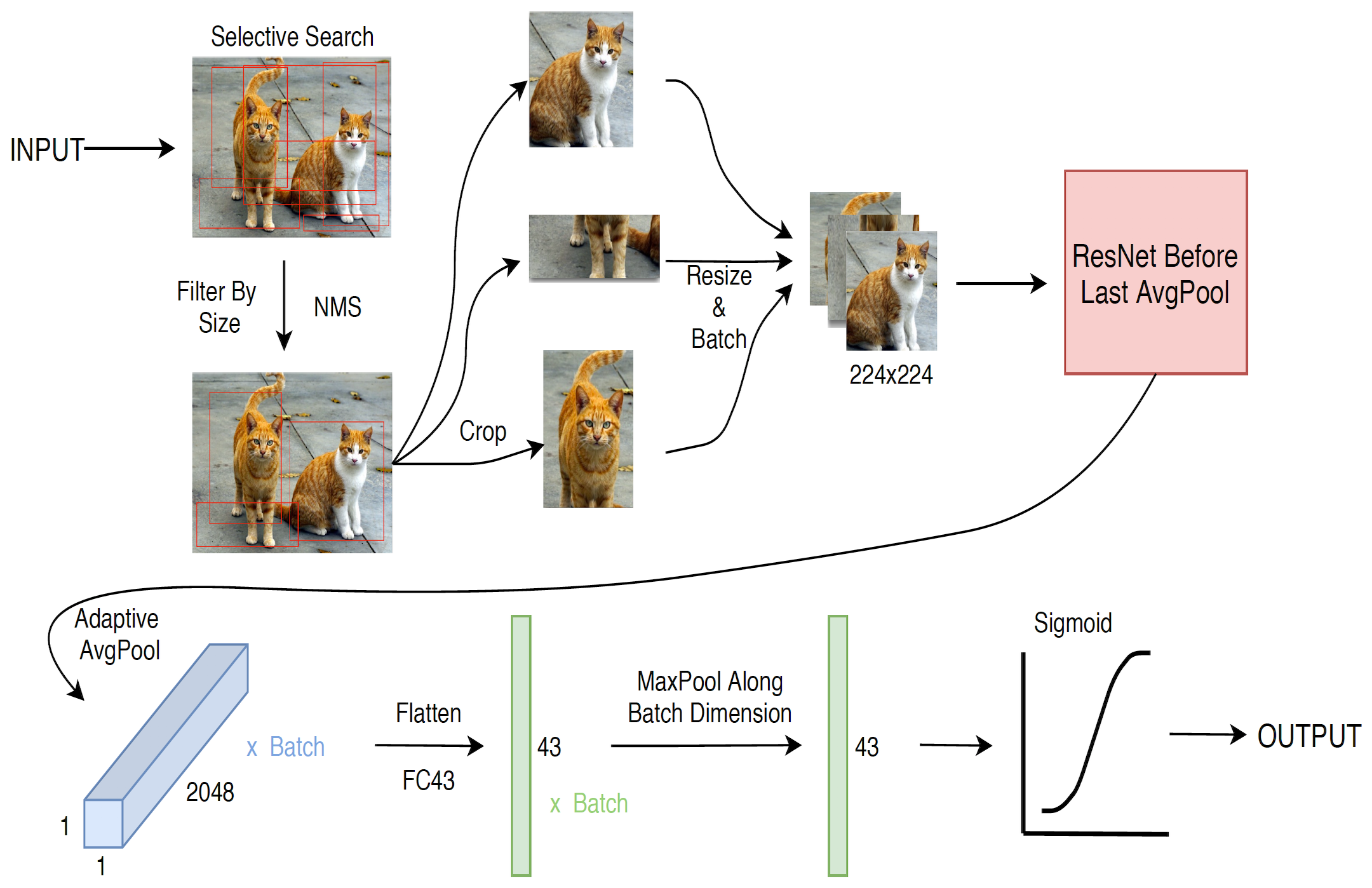}
\end{center}
   \caption{Architecture of proposed object proposal-based method.}
\label{fig:ss}
\end{figure}

Deep feature extractors are usually pre-trained on the single-label classification tasks, which are not suitable for classifying multiple objects in an image. In an attempt to solve this problem, we design a pipeline to transform the task of the feature extractor to single-label classification using selective search~\cite{uijlings2013selective}, which is an algorithm that generates multiple possible object locations in an image using graph-based segmentation techniques. The overview of the method is shown in Figure~\ref{fig:ss}.

We apply selective search at the beginning of the network to generate several bounding box proposals, some of which would ideally each contain one single object (actor) in an image. The proposal boxes are then filtered by their sizes, discarding those that have an area less than some threshold (we use 50 when testing). Then we implemented a non-maximum suppression~\cite{malisiewicz2011ensemble} to merge the boxes that have a large intersection over union (IoU) score (we use 0.7 when testing). The remaining box proposals are used to crop the parts of the original image and all the cropped images are resized to 224x224x3. Next, we use an additional dimension (e.g. batch) to make mini-batches of these N cropped images and pass them to the feature extractor (ResNet152 with an additional FC43 layer, as in baseline). After we get a score tensor of size Nx43x1, we do a max-pooling along the additional dimension and get one single vector of size 43x1. We then apply a sigmoid to it and get our normalized output prediction vector.

\subsection{Multi-Scale Fusion}

\begin{figure*}
\begin{center}
\includegraphics[width=0.8\linewidth]{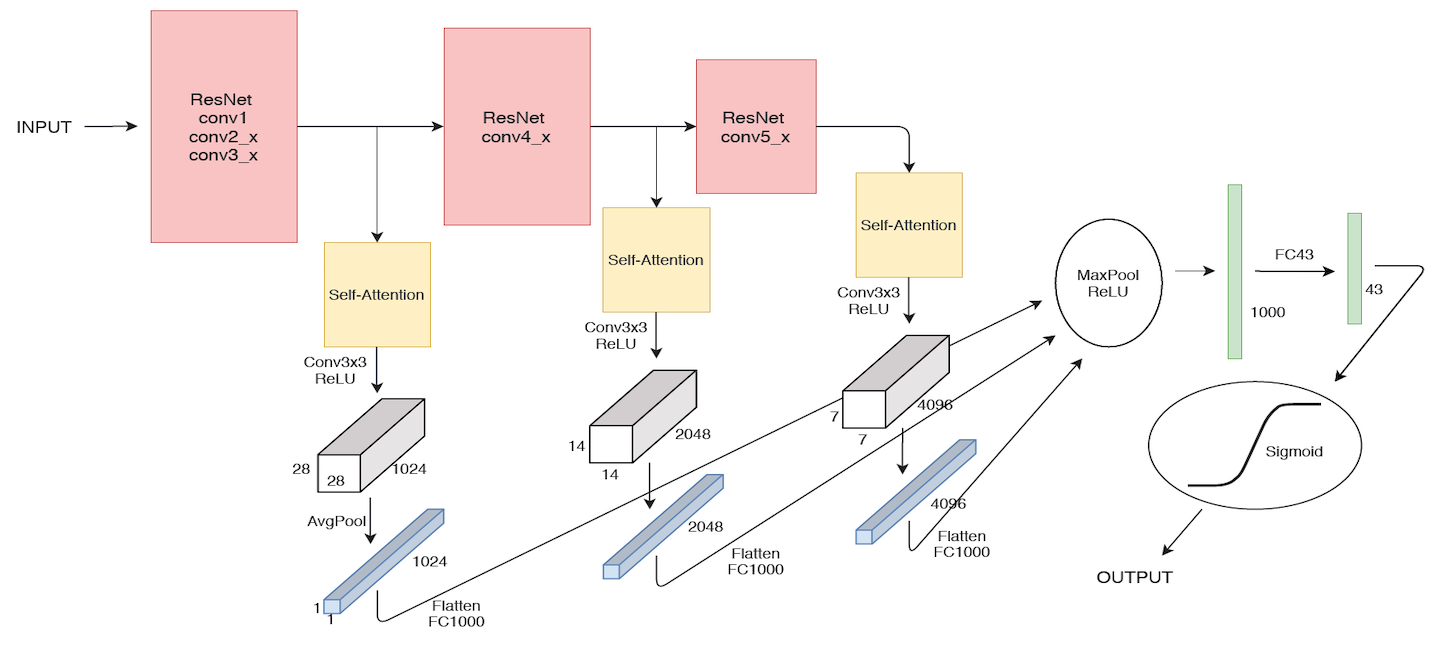}
\end{center}
   \caption{Architecture of proposed multi-scale fusion-based method.}
\label{fig:sa}
\end{figure*}

By observation, we notice some objects in many images in the dataset only occupy a small portion of the image. The ResNet architecture that we use as backbone introduces the idea of residual connections within four of its major blocks (named conv2\_x, conv3\_x, conv4\_x, conv5\_x in the original paper~\cite{he2016deep}). However, although the size of the image shrinks after passed through each block, there is no direct connection across these blocks. In order to reduce information loss, we add new branches after conv3\_x, conv4\_x, and conv5\_x to make individual predictions. The overview of the method is shown in Figure~\ref{fig:sa}. 

We add a self-attention module proposed in \cite{zhang2018self} to highlight the most useful information of the image at that scale. The output of the attention module is then passed to a 3$\times$3 convolutional layer and an adaptive average pooling layer to
size C$\times$1$\times$1, where the channel size C varies across different branches. After flattening and applying a fully connected layer, we retrieved three 1$\times$1000 vectors. We then do a max-pooling among the vectors to merge the predictions on different scales. We apply another fully connected layer of size 43 and then a sigmoid activation function to get the final prediction scores of 43 classes.

\section{Experiment}

\subsection{Training}

We use the same training configurations for all our models and train them until convergence. We build the models upon PyTorch deep learning framework\footnote{pytorch.org}. The backbone ResNet152 has been pretrained on ImageNet-1000~\cite{deng2009imagenet} and is further fine-tuned during training. We use the SGD optimizer with learning rate 1e-3, momentum 0.9, and weight decay 1e-4 to train the models. We use the per-class binary cross entropy (BCE) loss in our training. The weights of our proposal-based model converge at around 30 epochs and that of the fusion-based model converge at around 65 epochs.

\subsection{Results}

We train our models on the given A2D dataset containing 4750 training images. All three models use ResNet152 as their backbones, but it can be easily substituted with the less-powerful ResNet50 and ResNet101. The performance of our models on the released 1209 validation images is shown in Table~\ref{tab:performance}. Our object proposal-based method does not perform as good as we would expect, but our multi-scale fusion based method outperforms the baseline by a large amount.

\subsection{Discussion}

Results have shown that the utilization of information fusion on different scales and the object localization given by the self-attention module indeed increase the performance of the model. There are several possible reasons for the poor performance of the object proposal-based method. First, we merge the object proposals given by selective search simply by their IoU score without getting a confidence score. The merged boxes might not be the best ones for some objects. It might be a good idea to substitute the whole selective search and NMS part of the model by a pre-trained object detector such as the ones proposed in~\cite{bochkovskiy2020yolov4, ren2015faster}. Second, we use the batch dimension to group all the proposed bounding boxes, which prevents us from doing batch learning and significantly limits the speed of the training. Given this space for improvements, further work needs to be done before concluding that the object-level classification approach is not feasible.

\begin{table}
\begin{center}
\begin{tabular}{l|lll}
\hline
\textbf{Method} & \textbf{Precision} & \textbf{Recall} & \textbf{F1}\\
\hline
Baseline & 0.289 & 0.339 & 0.294 \\
Object Proposal & 0.241 & 0.303 & 0.253 \\
Multi-Scale Fusion & \textbf{0.521} & \textbf{0.554} & \textbf{0.515}\\
\hline
\end{tabular}
\end{center}
\caption{Performance of our models on the released validation set.}
\label{tab:performance}
\end{table}

\section{Conclusion}
We describe in detail the two different approaches we take for still image actor-action classification task on the A2D dataset. The proposed multi-scale fusion-based method significantly outperforms the object proposal-based method and the baseline method, achieving an F1 score of 51.5\% on the released validation set.

{\small
	\putbib[students/yzh232/egbib]
}

\end{bibunit}

\clearpage

\newpage

\setcounter{section}{0}
\setcounter{figure}{0}
\setcounter{equation}{0}
\setcounter{table}{0}
\title{Actor-Action Classification on the A2D Dataset: CSC249 Final Project}

\author{Ethan Fahnestock,  Bryce Ikeda,  Ian Lawson\\
University of Rochester\\
Hajim School of Engineering and Applied Sciences\\
Rochester, NY 14627\\
{\tt\small \{efahnest, bikeda, ilawson\}@u.rochester.edu} 
}

\maketitle

\begin{abstract}
  Actor and action labeling are often tackled separately. A unified approach for both of these tasks may leverage the relationship between an actor and its possible actions. We utilized YOLOv3 with actor thresholds set by particle swarm optimization (PSO) to perform actor classification. Then, this information was passed on to an action classifier based on ResNet-34 that took advantage of known actor-action relations to reduce the size of the action space. To train and verify our architecture, we use the A2D actor-action dataset and evaluate our predictions based on precision, recall and F1. Using our architecture on the validation set, we achieve a precision score of 42.0, recall score of 59.3 and F1 score of 46.3. 
\end{abstract}
\begin{bibunit}[unsrt]
\section{Introduction}
For this project we were required to develop a video actor-action classifier with a deep learning model. By training on an annotated video dataset, this model would learn to classify both the actors and their respective actions present in a video. This is a difficult problem due to the requirement of both object detection and action classification over a sequence of video frames. 

To train and test our model, we were provided with the A2D dataset. This is a dataset of 3782 YouTube videos, with 3036 videos designated for training and 746 for testing. Each video contains 3-5 annotated frames, where the annotations consist of actor and action labels. There are 7 possible actors (adult, baby, ball, bird, car, cat, dog) and 9 possible actions (climb, crawl, eat, fly, jump, roll, run, walk, none). In total, there are 43 possible actor-action pairs, as some actors may never be matched to a given action (adult-fly for example). 

To evaluate the performance of our model, we were required to test on three metrics: Precision, Recall and F1. Precision is defined as the number of true positives divided by the sum of true positives and false positives (or total predicted positives). Precision is composed of the rate of false positives. Recall is the number of true positives divided by the sum of true positives and false negatives (or total actual positives). Recall will be most affected by the rate of false negatives. F1 score is two times the precision times the recall divided by the sum of precision and recall. This is useful when it is important to avoid both false negatives and false positives equally \cite{acc_prec_rec_f1}.

\section{Related Works}
Much of contemporary research approaches the actor and action classification problem separately. For object detection, \cite{girshick2015fast} uses a Fast R-CNN but mistakes background patches in an image for objects because it does not understand the full context of an image. \cite{5255236} is a Deformable parts model which uses a sliding window approach filtering all positions and scales of an image. It uses bounding boxes for its training images and can detect highly variable objects. 

For action classification, \cite{6619185} uses spatiotemporal deformable part models by generalizing deformable part models from 2D images to 3D spatiotemporal volumes. Another action classification approach \cite{7410734} implicitly incorporates information about a human actor’s intentional movement. 

Recently, there has been research incorporating both the actor and action when classifying videos. Xu et al. \cite{Xu_2015_CVPR} approaches the actor-action segmentation problem by jointly inferring the actors and actions demonstrating the benefit of modeling the actor and action together. A weakly supervised actor-action segmentation is proposed by \cite{8099598} which shares information between different actors and actions. It learns a ranking matrix to select representative supervoxels for actors and actions and uses a conditional random fields model for semantic label inference. 

\section{Method}
Our group developed a method that splits actor-action classification. By separating the problem in this manner, the output space is reduced from 43 actor-action pairs to six for actor classification (combining adult and child, as explained later), and a maximum of thirteen actions. This comes at the cost of sequential dependence between actor-action classification. If actor classification fails, action classification will not be run, and the label will be incorrect. 

Our group was motivated to perform this split to simplify the classification problem and to reduce the data required for training, and training time. The actor and action classification performed is defined in the following sections.

\subsection{Actor Detection}
Actor classification was done using YOLOv3 \cite{redmon2018yolov3} pre-trained on the Coco 80 class dataset \cite{lin2014microsoft}. You Only Look Once (YOLO) is a real-time object detector that predicts bounding boxes as well as object classes. Its main feature is its ability to detect objects in one forward pass. YOLO uses a fully convolutional network including Darknet-53 as its feature extractor. Using these features, it predicts bounding boxes using dimension clusters as anchor boxes. Then, non-max suppression is used to reduce the number of bounding boxes by keeping only one box with the highest calculated objectness score for each ground truth object. This process is done across three different scales of the images to improve its resilience to small objects. 

YOLOv3 was run with its default configuration, a detection confidence threshold of 0.5, and an image size of 224x224. The COCO 80 dataset contains the classes person,sports ball, bird, car, cat, and dog, which map to the seven actor classes in the A2D dataset adult, child, ball, car, cat, and dog. Thus, detections of these six classes in the COCO dataset were used to determine if an actor was present in a video. 

\begin{figure}[H]
  \centering
  \includegraphics[width=0.45\textwidth]{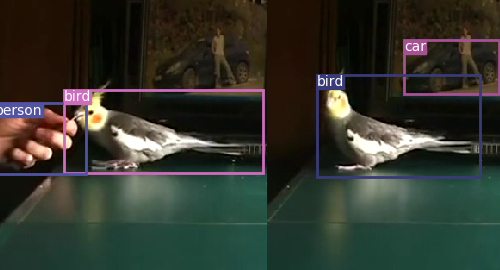}
  \caption{example of YOLO detections on A2D dataset}
  \label{fig:yolo_ex}
\end{figure}

An example of YOLO detections on two frames of a video in the A2D dataset can be seen in Figure \ref{fig:yolo_ex}. Notice in both frames that the bird is successfully detected. However, the person class is detected in the first frame, and the car class is detected in the second frame. This motivates the use of some sort of thresholding to eliminate detections of non-labeled actors. 

To detect actors as reliably as possible, YOLOv3 is run on the entire input video, and the number of actor class detections is divided by the number of frames in the input video, resulting in numbers between zero and one for each class, representing the percent of the video they were identified in. 

\begin{figure}[H]
  \centering
  \includegraphics[width=0.45\textwidth]{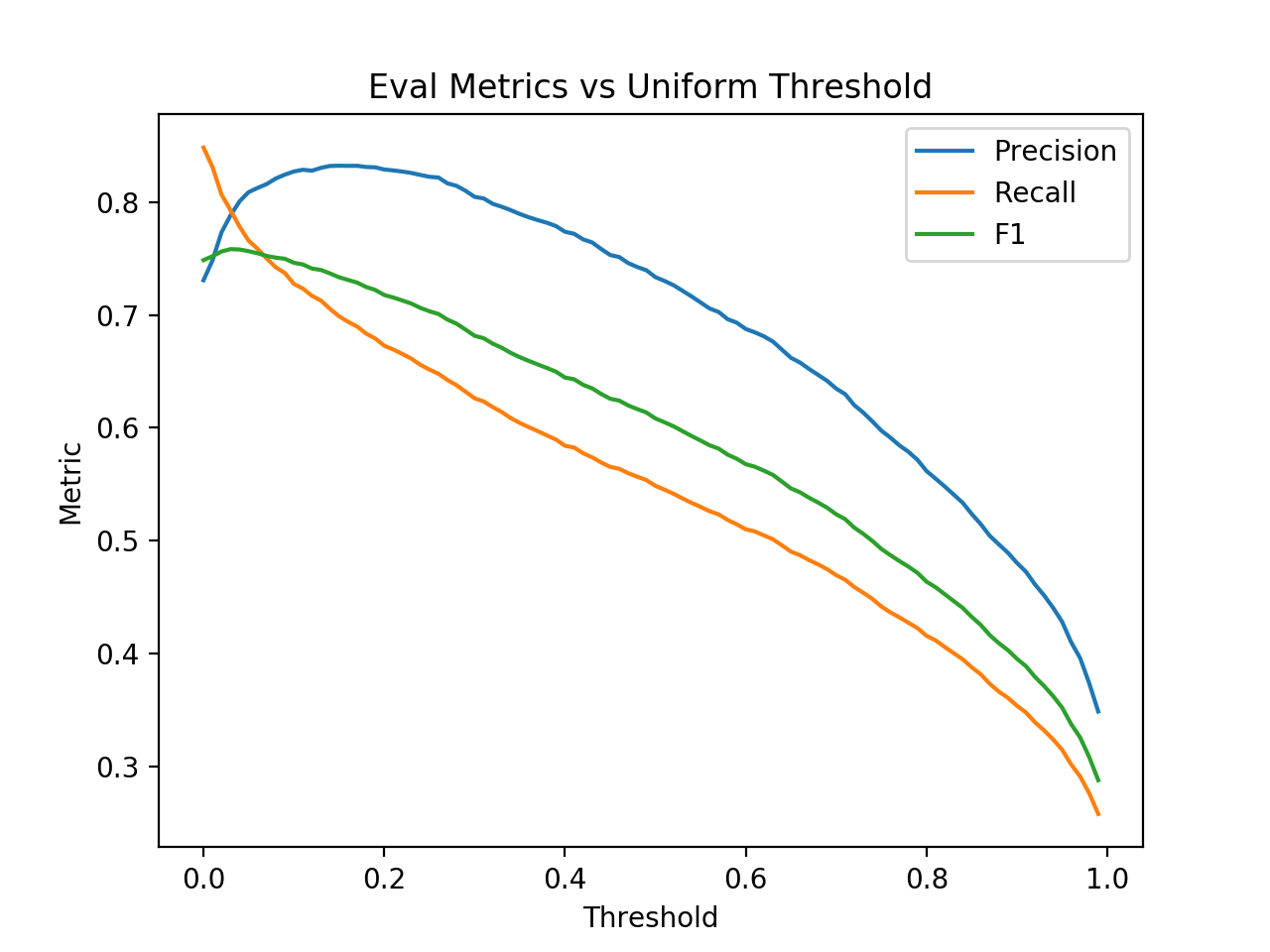}
  \caption{}
  \label{fig:eval_metrics}
\end{figure}

To select optimal threshold values to classify an actor as present/not present, the YOLOv3 detector was run on every video frame in the A2D dataset and detections were gathered by video. Video annotations were also gathered and compressed into all present actors in each video. This data was combined, and the evaluation metrics were calculated as the class threshold was varied between zero and one. This resulted in Figure \ref{fig:eval_metrics}.

Optimizing for precision, a uniform threshold for each class of 0.15 was selected. With this threshold, the actor detector had a precision of 83.2, recall of 69.9, and F1 of 73.4 on the entire annotated dataset (testing and validation). Data was not separated into testing and validation because no training was performed for actor classification. 

\subsubsection{Selection of Actor Thresholds with Particle Swarm Optimization}
As detection rates and accuracies vary across classes, we looked to further optimize the thresholds in a manner that they were unique for each class. After trying a few common optimization algorithms (BFGS, Nelder-Mead) we turned to Particle Swarm Optimization.

Particle Swarm Optimization (PSO) is a swarm-inspired optimization algorithm where particles navigate through a search space to optimize the provided fitness functions. The particles motion is controlled by ``dynamical equations'' with cognitive and social terms, which pull particles towards its neighbor's best previous position, and a particle's own best previous position respectively. Standard Particle Swarm Optimization (SPSO) \cite{zambrano2013standard} uses the following dynamical equations: 

\begin{equation}
  v_i(t+1) = wv_i(t) + x^{'}_i(t) -x_i(t)
\end{equation}

where $v_i(t)$ is the velocity of the $i$th particle at time $t$, $x_i(t)$ is its position, and $x_i^{'}$ is sampled from the hypersphere $\mathcal{H}_i(G_i, ||G_i - x_i||)$, with $G_i$ defined as:

\begin{equation}
  G_i = x_i + c\frac{p_i+l_i-2x_i}{3}
\end{equation}

with $c$ as a constant, $p_i$ the $i$th particle's previous best position, and $l_i$ as the previous best neighbor particle's position. 

We used an implementation of SPSO in python previously written by one of the group members. We ran SPSO with 40 particles and ten iterations per run, taking the best of four runs. Each dimension was bound between 0.1 and 0.4. Two simple fitness functions were used. First a fitness function of negative the precision value $-P$ was used (note the negative as the fitness function is minimized). This resulted in a high precision value, but at the cost of lowering recall and F1. The threshold vector generated also performed poorly when tested with the full action-actor network. Thus, a second fitness function multiplying the precision, recall, and F1 scores ($-P*R*F1$) was optimized, resulting in the threshold vector used in the final network ($[0.1,0.107,0.1,0.1,0.161,0.267]$ corresponding to [person,ball,bird,car,cat,dog]) which resulted in a final precision of 84.6, recall of 71.5, and F1 of 75.0 for actor detection. The metric performance for each of the thresholds discussed can be found in Table \ref{table:actor_performance}.

\begin{table}[H]
  \centering
  \begin{tabular}{@{}llll@{}}
  \toprule
  Threshold     & Precision & Recall & F1   \\ \midrule
  Constant 0.15  & 83.2        & 69.9     & 73.4   \\
  SPSO w/ -P  & 85.4        & 69.3     & 74.0   \\
  SPSO w/ -PRF1 & 84.6      & 71.5   & 75.0 \\
  \end{tabular}
  \caption{Actor Detector Performance With Different Thresholds}
  \label{table:actor_performance}
\end{table}

\subsection{Action Detection}
For action detection we trained ResNet networks initialized with pre-trained weights for each actor. Residual Networks (ResNet) \cite{he2016deep} aim to solve the degradation problem in deep learning applications. The degradation problem is when a network reaches a high enough number of layers, the training error begins to rise due to vanishing gradients. To tackle this problem, residual learning is applied. Residual learning is done through a combination of residual blocks. Residual blocks use skip connections to pass its intermediate activations deeper into the neural network. This simple passing of information, in tandem with multiple residual blocks allow ResNet to train deeper neural networks without degrading the network.

Only annotated video frames were used in training. The number of data points used to train and validate each actor class can be found in Table \ref{tab:actor_data}. Several decisions were made while optimizing network performance, which are detained in the following section. 

\begin{table}[H]
  \centering
  \begin{tabular}{@{}lll@{}}
  \toprule
  Class  & Num Training & Num Val \\ \midrule
  Person & 1397         & 408     \\
  Ball   & 370          & 101     \\
  Bird   & 757          & 213     \\
  Car    & 599          & 162     \\
  Cat    & 738          & 199     \\
  Dog    & 889          & 226    
  \end{tabular}
  \caption{training and validation counts by actor}
  \label{tab:actor_data}
\end{table}

\subsubsection{Choice of Network Hyperparameters and Training Method}

Several network configurations were tested and their results were analyzed to optimize network performance. Note that this section describes the methods that were applied to arrive at the utilized architecture. The ranking of methods was based on limited tests used to make design decisions, and we do not claim that the patterns we saw are generalizable properties. 

\subsubsection*{Activation Function}
ReLU and Sigmoid activation functions were both tested for acting on the final layer. On several examples the sigmoid activation function was observed to result in better precision, recall, and F1 values when the trained model was evaluated on the validation dataset. Thus, sigmoid was chosen as the activation function acting on the final layer. ReLU was initially chosen and was used in the rest of the network because it is computationally more efficient. The performance of ReLU and Sigmoid activation functions on the validation dataset for each class can be seen in Tables \ref{tab:relu_results} and \ref{tab:sigmoid_results} respectively.

\begin{table}[H]
  \centering
  \begin{tabular}{@{}llll@{}}
  \toprule
  Class Name & Precision & Recall & F1   \\ \midrule
  Person     & 28.3      & 59.0   & 35.4 \\
  Ball       & 38.1      & 49.0   & 35.4 \\
  Bird       & 35.8      & 74.8   & 46.1 \\
  Car        & 36.0      & 94.8   & 51.1 \\
  Cat        & 32.9      & 59.5   & 39.8 \\
  Dog        & 31.7      & 56.4   & 39.6 \\ \bottomrule
  \end{tabular}
  \caption{Action Classifier Performance on the Eval Dataset Using ReLU, 45 epochs}
  \label{tab:relu_results}
\end{table}

\begin{table}[H]
  \centering
  \begin{tabular}{@{}llll@{}}
  \toprule
  Class Name & Precision & Recall & F1   \\ \midrule
  Person     & 38.2      & 80.4   & 49.4 \\
  Ball       & 49.0      & 77.2   & 58.3 \\
  Bird       & 46.8      & 85.3   & 57.6 \\
  Car        & 56.8      & 70.7   & 61.3 \\
  Cat        & 36.1      & 65.1   & 44.4 \\
  Dog        & 36.7      & 75.2   & 47.5 \\ \bottomrule
  \end{tabular}
  \caption{Action Classifier Performance on the Eval Dataset Using Sigmoid, 45 epochs}
  \label{tab:sigmoid_results}
\end{table}

\subsubsection*{Training Epochs}
\begin{figure}[H]
  \centering
  \includegraphics[width=0.45\textwidth]{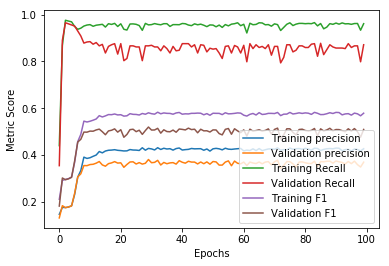}
  \caption{training and validation metric scores vs epoch}
  \label{fig:epoch_vs_learning}
\end{figure}

Several epoch lengths were tested, ranging from 10-100 epochs. As can be seen in \ref{fig:epoch_vs_learning}, performance metrics plateau after only 10-15 epochs, after which over-fitting begins on the training dataset. Based on this data, a training duration of 15 epochs was selected. 

\subsubsection*{Loss Functions}

Two loss functions were tested as well. As the problem posed, even when decoupled into independent actor and action classification, is a multi-label multi-classification problem, Pytorch's BCEWithLogitLoss was first used as it is designed to work with multi-label classification problems. After testing different epoch lengths and network configurations with both BCEWithLogitLoss and BCELoss, performance metrics above 15\% were not achieved, and the loss function was transitioned to single class Cross Entropy Loss.

\subsubsection*{Size of Resnet}

\begin{figure}[H]
  \centering
  \includegraphics[width=0.45\textwidth]{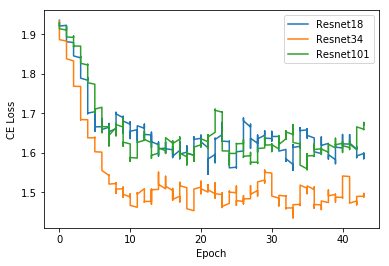}
  \caption{median filtered loss functions across tested Resnet size}
  \label{fig:resnet_compare}
\end{figure}

Resnet18, Resnet34, and Resnet101 were all tried as the base Resnet in the action detection network. Resnet34 outperformed the other two in loss as can be seen in Figure \ref{fig:resnet_compare}. Table \ref{tab:resnet_changes_metrics} shows the performance metrics for the three networks trained using varied Resnet depth for bird action classification. Resnet34 was selected over Resnet18 for its performance on minimizing the loss, small training time penalty over Resnet18, and similar precision performance to Resnet18. 

\begin{table}[H]
  \centering
  \begin{tabular}{@{}lllll@{}}
  \toprule
  Resnet     & Precision & Recall & F1   & Train Time (s)\\ \midrule
  Resnet 18  & 41        & 89     & 54   & 794.92\\
  Resnet 34  & 41        & 85     & 57   & 1070.43\\
  Resnet 101 & 35.7      & 87.8   & 49.9 & 2326.60\\
  \end{tabular}
  \caption{Resnet performance on classifying bird actions, 45 epochs, w/3 FC layers}
  \label{tab:resnet_changes_metrics}
\end{table}

\subsubsection*{Additional FC Layers}

Initially, the network included two fully-connected (FC) layers fed by the FC output of Resnet. The first of these layers had 40 neurons, and the second and final layer had one neuron per valid class action, varying between 4 to 13 between actor classes. ReLU activations were placed between network layers. The network was tested without the additional FC layers as well and had noticeably better precision performance at the cost of losing precision performance. As precision was the focus of this project, the network configuration without the additional FC layers was selected. The performance metrics on the trained networks averaged over actor classes with and without the additional FC layers can be seen in Table \ref{tab:fc_layer_results}.

\begin{table}[H]
  \centering
  \begin{tabular}{@{}llll@{}}
  \toprule
  Network Config     & Avg. Precision & Avg. Recall & Avg. F1  \\ \midrule
  With FC Layers  & 42.7        & 77.1     & 52.4   \\
  W/o FC Layers  & 56.9        & 65.9     & 59.2   \\
  \end{tabular}
  \caption{average action classifier performance}
  \label{tab:fc_layer_results}
\end{table}

\begin{figure}[H]
  \centering
  \includegraphics[width=0.4\textwidth]{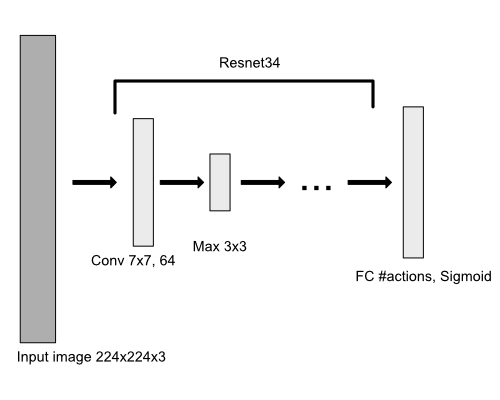}
  \caption{final action classification network}
  \label{fig:action_network}
\end{figure}

With this decision made, the actor class specific action classification networks took the form used for this project, which is summarized in Figure \ref{fig:action_network}.

\subsubsection*{Remaining Training Parameters}
Stochastic gradient dissent was used with a learning rate $lr=0.001$ and momentum of $0.9$. A learning rate scheduler was used to decay the learning rate every 7 epochs to 10\% of its previous value. Input data random cropping was disabled, along with flipping. When annotated images with more than one classified action was provided the first present action after the first one was selected. The model weights were saved every 10 steps and every time a new best loss value was achieved. A batch size of 4 was used.

\begin{figure*}
  \centering
  \includegraphics[width=0.9\textwidth]{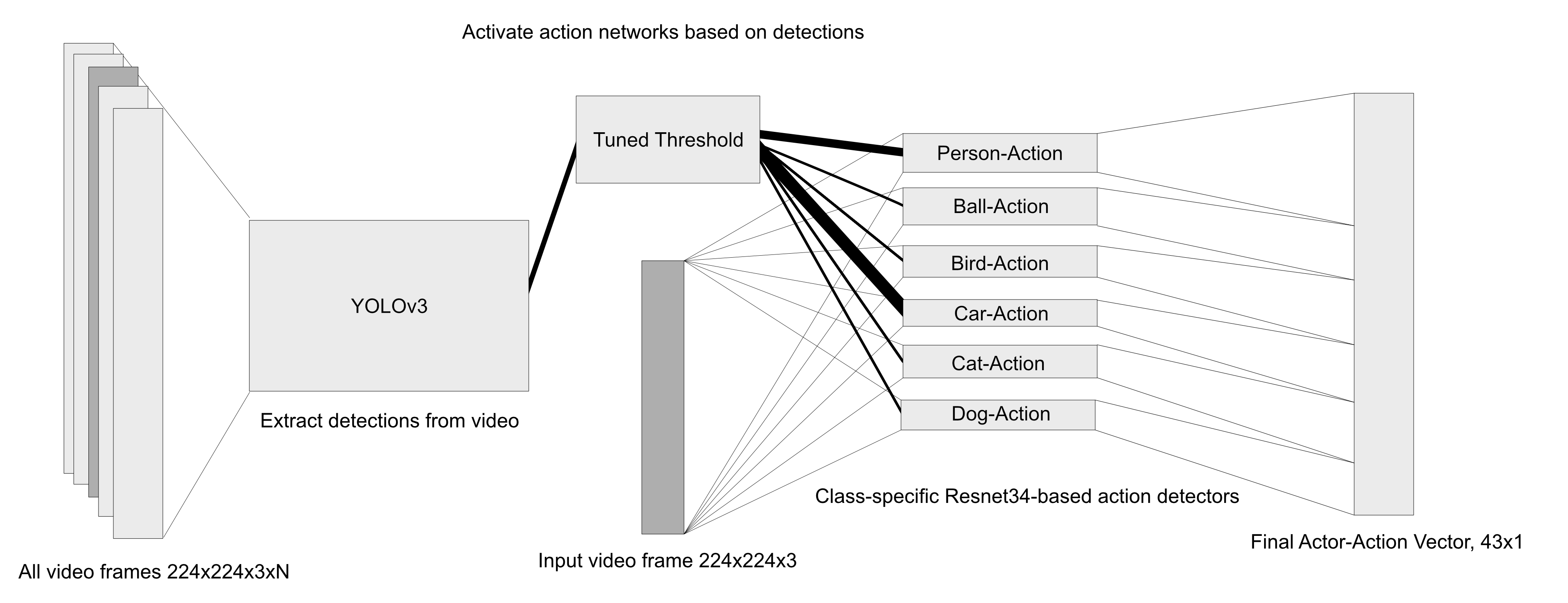}
  \caption{final action-actor classification network layout}
  \label{fig:full_network}
\end{figure*}

\subsubsection*{Final Training Statistics and Network Structure}
The final high-level network structure can be seen in Figure \ref{fig:full_network}. The final metric performance of the class-specific action classifiers and training time can be found in Table \ref{tab:final_training_stats}. This network was trained on a Google Cloud Deep Learning VM with 2 cores, 13GB of ram, and a NVIDIA Tesla K80.

\begin{table}[H]
  \centering
  \begin{tabular}{@{}lllll@{}}
  \toprule
  Class  & Precision & Recall & F1   & Training Time (min) \\ \midrule
  Person & 57.7      & 80.0   & 63.7 & 67.9                \\
  Ball   & 61.9      & 62.9   & 61.9 & 9.1                 \\
  Bird   & 55.0      & 65.3   & 63.4 & 25.25               \\
  Car    & 63.6      & 64.8   & 63.4 & 17.8                \\
  Cat    & 53.0      & 61.6   & 55.0 & 33.12               \\
  Dog    & 50.3      & 60.5   & 53.2 & 36.9                \\ 
  Total  &           &        &      & 190.1                \\ \midrule                                        
  \end{tabular}
  \caption{final training/performance statistics}
  \label{tab:final_training_stats}
\end{table}

\section{Benefits of Network Structure}
This approach selects to decompose the provided problem over learning an end-to-end approach. By doing so this method exposes parameters allowing the tuning of its behavior on a level not directly achievable from an end-to-end solution for action-actor classification. Specifically, the thresholds used to activate the sub-network action classifiers can be tuned by a human for their application. 

Our use of pre-trained networks enables our solution to learn new actor classes quickly, with many fewer annotated examples than an end-to-end deep learning solution would require. An average of 800 annotated action examples was used for each class. 

The modularity of this network also allows the easy addition or removal of an actor class without the need to retrain any part of the network. If better classification of a specific actor's actions is desired, or a new action is to be added, only that sub-network must be retrained.

\section{Performance on Validation Set}
\begin{figure}
  \centering
  \includegraphics[width=0.4\textwidth]{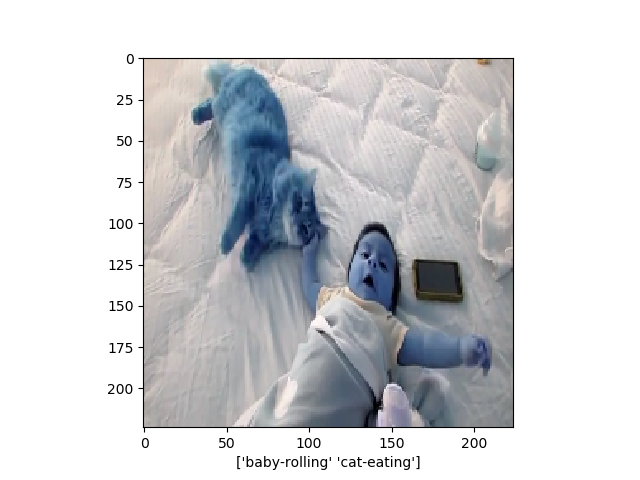}
  \caption{example of generated actor-action pairs}
  \label{fig:examples}
\end{figure}

Our method's final performance on the validation dataset can be seen in Table \ref{tab:final_performance}. Figure \ref{fig:examples} shows randomly picked images from the test dataset and have their predicted captions. In general it appears the actor classification works well, and the action classification tends to be generous. 
\begin{table}[H]
  \centering
  \begin{tabular}{@{}lll@{}}
  \toprule
    Precision & Recall & F1  \\ \midrule
    42.0 & 59.3 & 46.3 \\                                  
  \end{tabular}
  \caption{final performance of whole network}
  \label{tab:final_performance}
\end{table}

\section{Conclusion}
In this project we approached actor-action classification. Our approach split actor and action classification, but leveraged information on actor-action relationships in the A2D dataset to inform action classification. Our network used YOLOv3 for actor detection and ResNet-34 for action classification. We tested our network on the evaluation network using Precision, Recall and F1 score. Future work could include exploring the possibility of using the bounding boxes in YOLOv3 to do cropping for action detection. Another improvement could be to use YOLOv3 only on frames close to the requested frame.

\section*{Links to Utilized Resources}
\vspace{-0.1in}
\begin{itemize}[noitemsep,topsep=0pt]
  \item \href{https://github.com/pathak22/pyflow}{pyflow} was used to generate the optical flow results
  \item \href{https://github.com/eriklindernoren/PyTorch-YOLOv3}{PyTorch-YOLOv3} was used for all YOLO classification
\end{itemize}

{\small
\putbib[students/efahnest/ref]
}

\end{bibunit}
\clearpage

\newpage
\setcounter{section}{0}
\setcounter{figure}{0}
\setcounter{equation}{0}
\setcounter{table}{0}
\title{Transfer Learning for Multi-Label Actor-Action Classification}

\author{Alan Finkelstein\\
Department of Biomedical Engineering, University of Rochester\\
\\
{\tt\small alan$\_$finkelstein@urmc.rochester.edu}
}

\maketitle

\begin{abstract}
    Here we present a deep learning model, using transfer learning, for multi-label actor-action classification in the A2D dataset. Transfer learning, using ResNet-152 was implemented, experimenting with various hyperparameters and frozen layers. The best performance was achieved using Binary cross entropy loss with logits , Adam optimizer, with a learning rate of 0.0001. No weight decay was applied. Using our model, and training for 50 epochs we achieved a precision score of 49.3, a recall of 57.9 and an F1-score of 50.5 on the validation set. 
\end{abstract}

\begin{bibunit}[unsrt]
\section{Introduction}

Multi-label actor-action classification is a challenging task in computer vision. Variability in image quality, image intensity, and apparent optical flow due to camera motion complicate standardization of this problem. If actor and action are coupled, overfitting can also occur due to class imbalances. Another common challenge when training a classification model is the limited size of datasets, which can lead to poor network performance. Transfer learning is a suitable option to address this problem by transferring the weights of a pretrained model to a new network. In this way low-level features learned from a previous task can improve the accuracy of the model and result in quicker convergence. In our method we implemented transfer learning to solve the multi-label actor-action classification problem utilizing ResNet-152 as a base model. 

\section{Methodology}

\subsection{Data}
In this project we utilized the A2D dataset, which contains 3782 videos from YouTube. Single frames were extracted from each video, with objects annotated with an actor-action label in each frame. Each frame was processed and loaded using with a batch size of 32 for the training and validations sets. Images were read, converted to tensors in pytorch and normalized. The processed images (224x224x3) and their class labels (43-D encoding) were returned and used for training and validation, respectively. The testing dataset was processed similarly, without annotation labels.

\section{Experiments}

\subsection{Transfer Learning}
Our model was built using ResNet-152 as the base model, pretrained. ResNet 152 is a very deep network, comprised of skip connections that allow the network to fit the input from the previous layer to the next layer without any modification of the input. Winner of ILSVRC 2015, ResNet-152 is capable of image classification, detection, localization, and segmentation~\cite{he2016deep, tan2018survey}. Performance using a pretrained network is only as good as the similarity between the datasets used for pretraining and fine tuning. Experimentation with freezing layers is essential to determine the optimal model architecture to achieve improved network performance and convergence. If all of the layers are unfrozen, so that all parameters are updated with each batch, then the model may take too long to converge. Freezing too many layers prevents the model from learning features specific to the new dataset. Course features, such as edges, are usually learned by early layers, and so no new information is gained by updating the parameters of those layers. Nonetheless, we experimented by freezing layers 1-6 and evaluated the impact of freezing layers on model performance. Lastly, the final linear (fully connected) layer was changed to be consistent with the number of labels in our dataset (43).

\subsection{Hyperparameters}
\subsubsection{Loss Function}
To train our network, binary cross entropy with logits, or binary cross entropy (BCE) with sigmoid was used as the loss function. BCE with logits is analogous to cross entropy loss, which will implicitly apply softmax activation. Binary cross-entropy loss is a sigmoid activation plus a cross-entropy loss. BCE is independent for each vector component or class, that is the loss is computed independently for every CNN output vector, and is not affected by other component values~\cite{buja2005loss}. This attribute makes BCE loss well suited for multi-label classification~\cite{liu2017deep}. That is, the probability of any element belonging to one class is independent of the decision, or probability, of that same element, or another belonging to a different class.

\subsubsection{Optimizer}
Adam was used as the optimizer, with a learning rate of 0.0001 to train the network. Adam is an adaptive learning rate optimization designed for deep neural networks. Individual learning rates are found for each parameter, which helps with convergence~\cite{kingma2014adam}{}. Various learning rates were tried to optimize our network performance, as shown in our results. Various weight decays (l2 regularization) were also implemented, but adding an l2 regularization did not improve network performance.

\begin{figure}[H]
    \centering
    \includegraphics[width=1.0\linewidth]{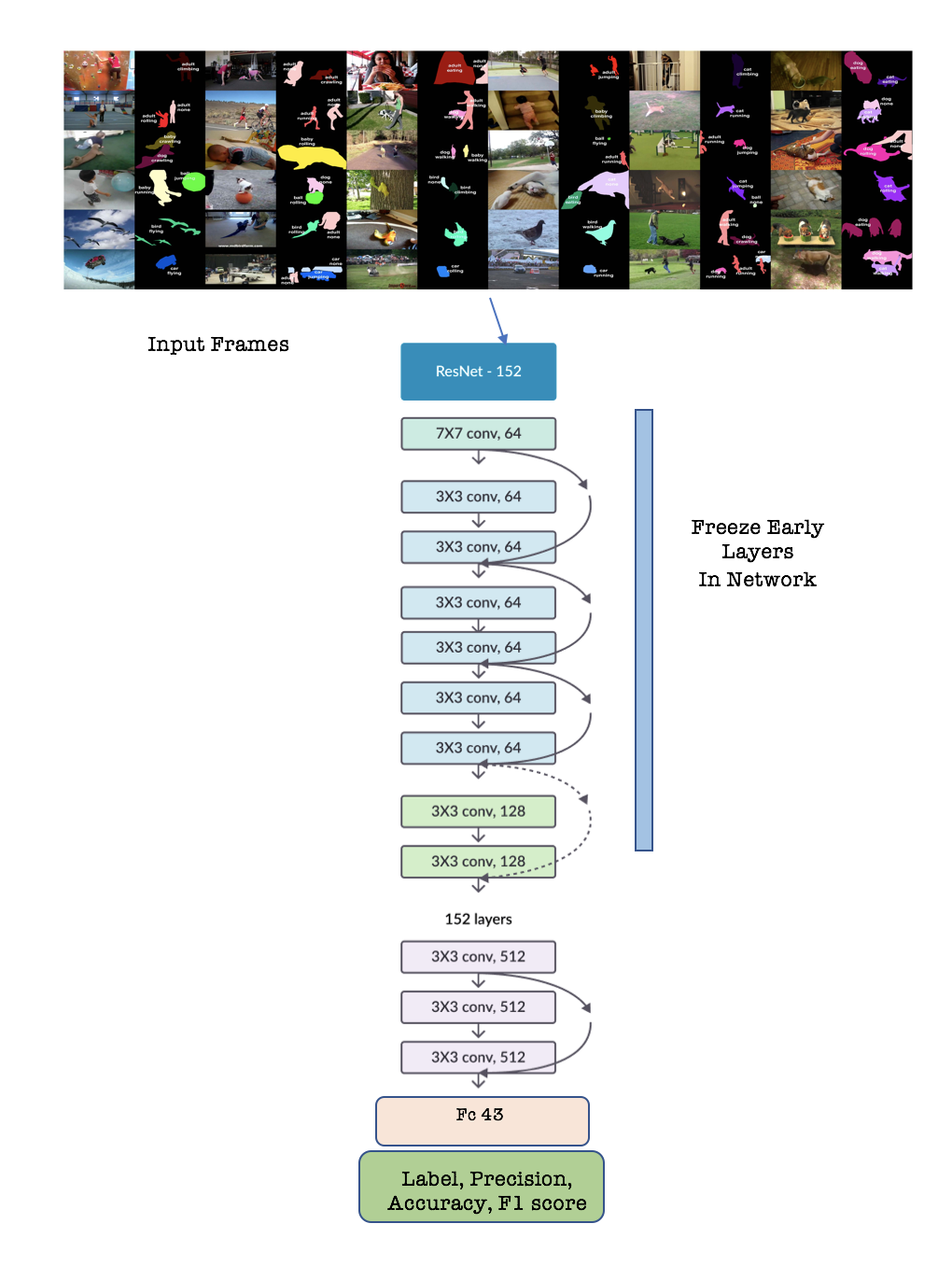}
    \caption{\textbf{Model architecture for actor-action (A2D) multi-label classification.} ResNet-152 was used as the base model with layers 1-6 frozen. Network parameters for layers 7-9 were updated during training. A final fully connected layer with 43 classes was used.}
    \label{fig:my_label}
\end{figure}

\section{Results}
\subsection{Frozen Layers}
\begin{figure}[H]
    \centering
    \includegraphics[width=1.0\linewidth]{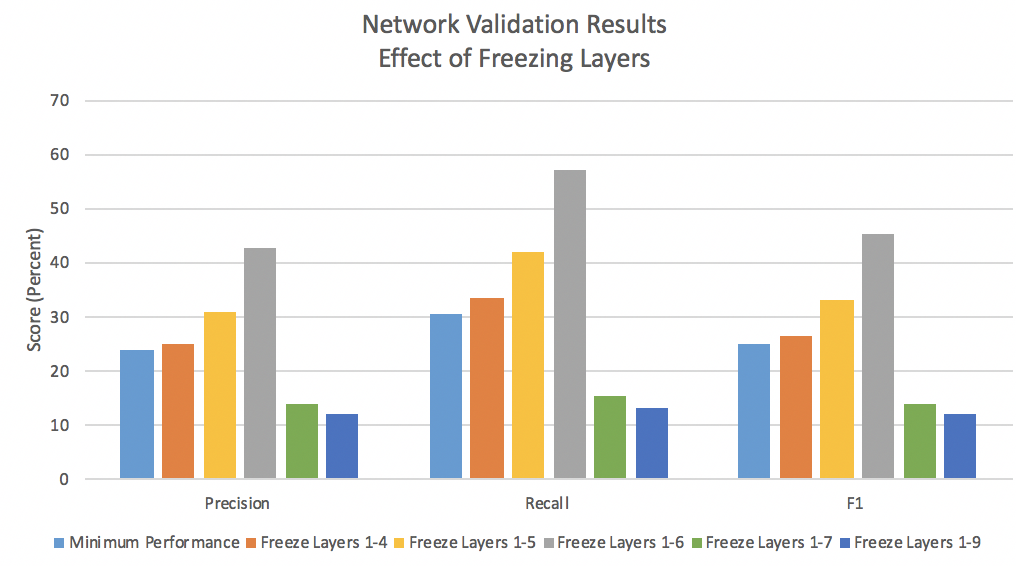}
    \caption{\textbf{Effect of freezing layers in pretrained network.} Frames are fed into ResNet-152 pretrained network. Layers 1-6 were frozen, and last fully connected layer was changed to 43 to be consistent with target size.}
    \label{fig:my_label}
\end{figure}

\subsection{Learning Rate}
\begin{figure}[H]
    \centering
    \includegraphics[width=1.0\linewidth]{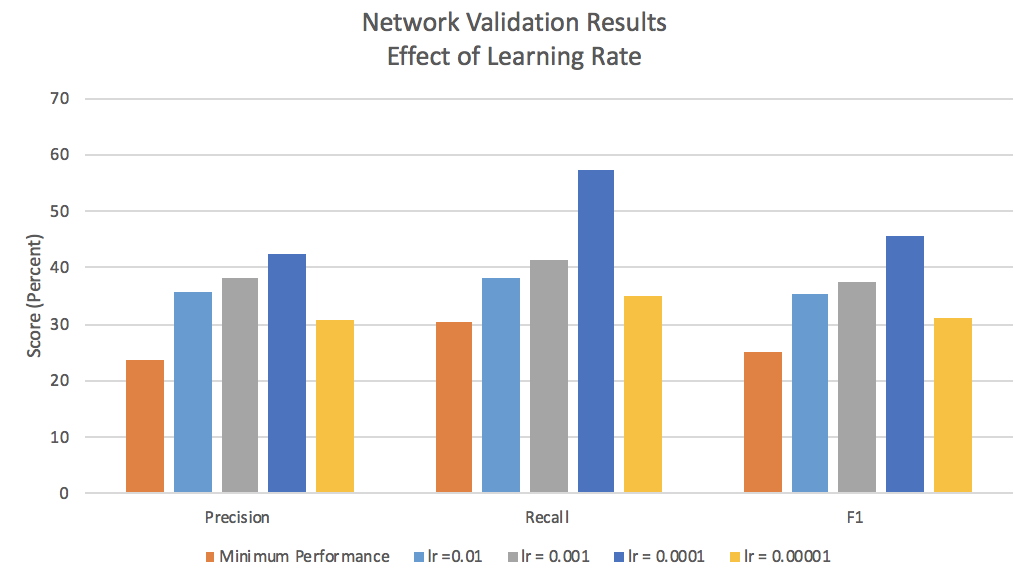}
    \caption{\textbf{Effect of learning rate on network performance.}}
    \label{fig:my_label}
\end{figure}

The effect of learning rate was assessed holding all other hyperparameters constant. That is models were trained using BCEWithLogitsLoss and Adam for 10 epochs while varying the learning rate. Adam trains quickest when using a learning rate of approximately 0.0005 to 0.01~\cite{kingma2014adam}. This was confirmed in our results, which showed that a lr of 0.0001 outperformed all other models for the same number of epochs.

\subsection{Number of Epochs}

\begin{figure}[H]
    \centering
    \includegraphics[width=1.0\linewidth]{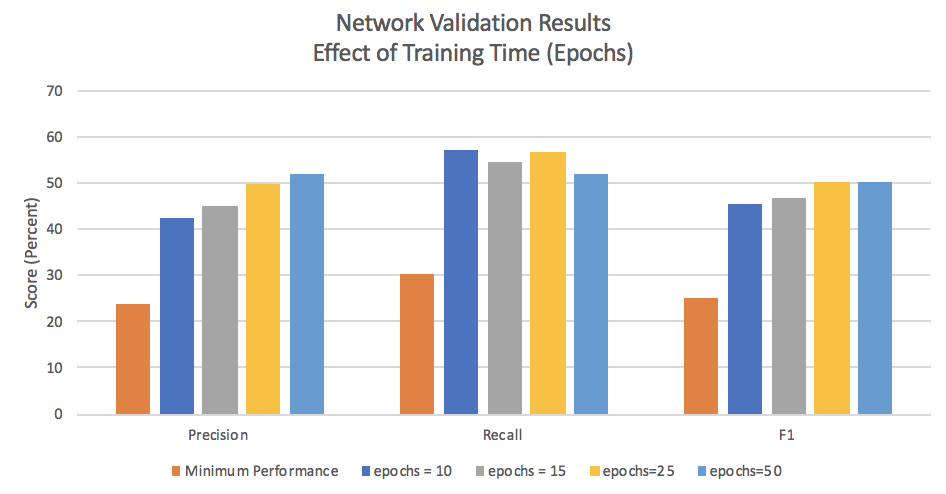}
    \caption{\textbf{Effect of Training time (epochs) on network performance.}}
    \label{fig:my_label}
\end{figure}

Training for a more epochs improved the precision and F1 score however worse recall was observed with more epochs. Depending on the application a choice could be made to opt for better precision or recall. 

\subsection{Optimizer}
\begin{figure}[H]
    \centering
    \includegraphics[width=1.0\linewidth]{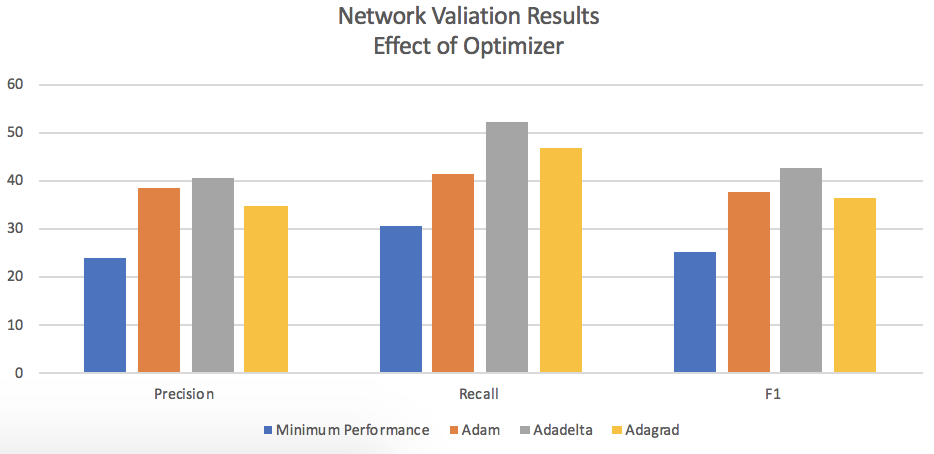}
    \caption{\textbf{Effect of different optimizers on network performance.}}
    \label{fig:my_label}
\end{figure}

A comparison of different optimizers with default parameters on network performed revealed that models which dynamically updated the learning rate performed better (such as Adadelta and Adagrad). However hyperparameter tuning of these optimizers was not performed. Adadelta was first proposed in ADADELTA: An Adaptive Learning Rate Method~\cite{zeiler2012adadelta}. The model dynamically adapts over time and requires no manual tuning of the learning rate. Adagrad was proposed in 2011~\cite{duchi2011adaptive}. Adagrad employs an adaptive method that initiates the learning rate and applies an adapted learning rate decay component wise at each step.

\section{Conclusions}

For this project, we proposed a deep learning model for multi-label classification of the A2D actor-action dataset. Our approach incorporated a pretrained network, ResNet-152, as the base model. Freezing the first 6 layers, and training the rest appeared to provide the best network performance, and offered the best trade off given the size of our dataset. Training for more epochs did improve network performance but a significant difference was not observed beyond 25 epochs. While overfitting is always a concern, it did not seem to be an issue when evaluating our performance metrics on the validation dataset. However in future applications our model could be trained on a larger dataset, or for longer amounts of time incorporating a regularization parameter, such as weight decay (l2 regularization) to prevent overfitting.  Additionally, implementation of various data augmentation approaches, such as rotation, cropping, and horizontal flip, would help to artificially increase the size of the training dataset~\cite{perez2017effectiveness}.

While the proposed model and training paradigm offered acceptable results, much could be done to improve future applications. Class imbalance was not addressed in this project, but is certainly an issue that should be addressed. Coupling of the actor-action classification results in a class imbalance since specific cases of a specific actor-action may only occur a couple of times. Decoupling the classification into two tasks, actor classification and action classification would help to resolve this and potentially improve the performance and reliability of the model. Additionally only single frames were used from each YouTube video in the dataset. Use of video features, such as more frames or optical flow information would improve the action recognition task in future applications. Additionally other CNN architectures have been shown to exhibit better performance, such as RNN-CNNs, and CNNs coupled with attention mechanisms ~\cite{wang2017residual,guo2018cnn}.

{\small
\putbib[students/writeup1/bib]
}

\end{bibunit}
\clearpage

\newpage

\emptythanks

\setcounter{section}{0}
\setcounter{figure}{0}
\setcounter{equation}{0}
\setcounter{table}{0}

\title{Weighted Multi-Label Actor-Action Classification}

\author{Kehao Guo\\
University of Rochester\\
{\tt\small kguo@u.rochester.edu}
}

\maketitle

\begin{abstract}
	In this report, I describes technical details of my attempt to achieve multi-label actor-action classification based on the A2D dataset. The model I proposed is based on a pretrained ResNet-152 model with additional custom layers and a feature extractor. My approach also involves applying positive class weights to address imbalance data. My final submission achieves a F1 score of 34.6.
 \end{abstract}

\begin{bibunit}[unsrt]

\section{Introduction}

Object classification and activity recognition are well-developed techniques supported by abundant algorithms and datasets. However, the common disregard for types of actors when inferring actions tended to cause invalid and illogical predictions on actor-action pairs. The A2D dataset was introduced to address this problem by jointly considering various types of actors undergoing various actions. It contains images as labelled frames of trimmed video clips from Youtube that describe 43 valid actor-action pairs from 7 classes of actors and 9 classes of actions such as "cat-climbing" and "dog-running". There are in total 4750 training images, 1209 validation images and 1044 testing images \cite{a2d}.

Based on the dataset, I presented a model that use a pretrained ResNet-152 model as a backbone with weights added to positive classes during training to improve validation accuracy. The model was able to achieve a F1-score of 34.6 during testing.

\section{Method}

\subsection{Data processing}

I used a dataloader provided in \cite{git} for processing training and validation sets. It takes images and annotations as inputs and outputs the processed images with a dimension of 224*224*3 and their class labels with 43-D encoding represented by 1s and 0s. The dataloader also employs data augmentation techniques (such as padding, cropping, flipping and rotation), normalization and shuffling to enhance the diversity of training data.

\subsection{Model design}

\begin{figure}[t]
\begin{center}
\includegraphics[width=0.8\linewidth]{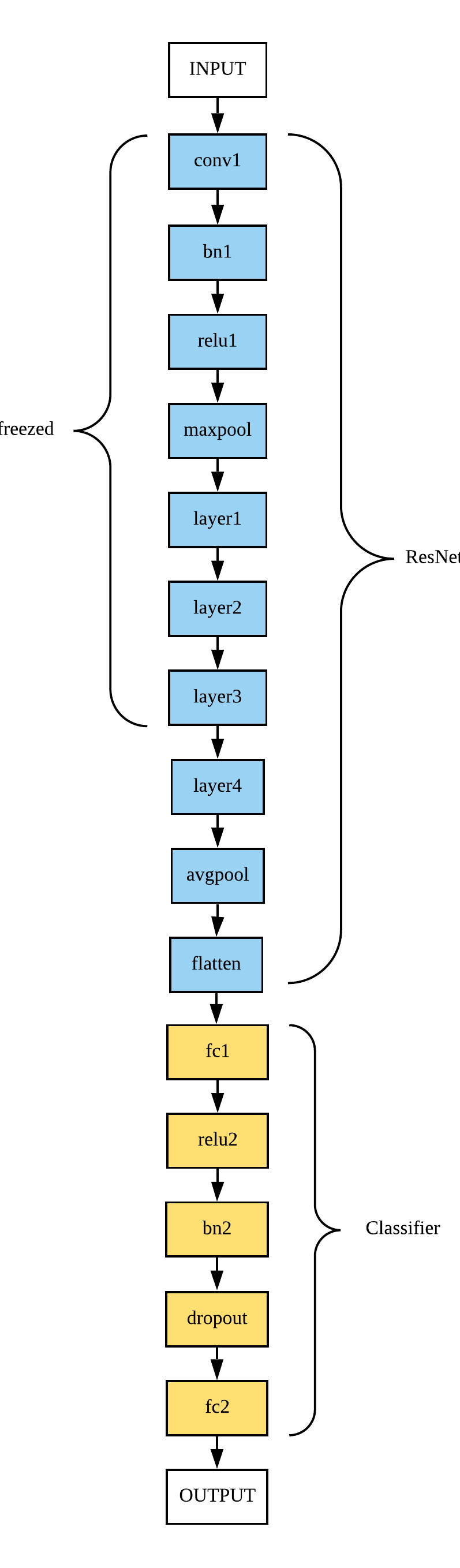}
\end{center}
   \caption{Network structure}
\label{fig:p1}
\end{figure}

I used the pretrained ResNet-152 model as a feature extractor by freezing layers in the top
modules which capture relatively general and abstract features in the image. I finetuned the
rest of RestNet-152 (the last 2 modules: layer4 and avgpool) so that they can capture
features that are specific to samples in A2D dataset. Finally, I replace the fully connected layer in
RestNet-152 with a customized classifier for feature extraction. The structure of the model is shown in figure \ref{fig:p1}. The last fully connected layer has an output
size of 43, which is the number of classes.

Criterion used is BCEWithLogitsLoss, which is the combination of a sigmoid layer and binary
cross entropy loss. The loss is defined as \cite{bce}:
\begin{center}
    \begin{equation}
        \mathit{l(x,y)} = mean\{l_1, ..., l_N\}^T,
    \end{equation}
    \begin{equation}
        \mathit{l_n} = -w_n [y_n\cdot\log\delta(x_n)+(1-y_n)\cdot\log(1-\delta(x_n))],
    \end{equation}
\end{center}
where \textit{N} is the batch size.

Optimization method used is Adam. Initial learning rate is set to 0.0005. I adopt an adaptive
learning rate schedule. Specifically, I reduce learning rate by a factor of 4 when running
loss stops improving for at least 5 epochs.

Other hyperparameters include:
\begin{itemize}
  \item Mini-batch size: 4
  \item Weight decay: 0
  \item Epochs: 50
\end{itemize}

\subsection{Adding class weights}

I implemented and applied class weight to the loss function. This deals with the problem of unbalanced data in order to improve validation performance. Since there exists only a smaller number of actor-action pairs in an
image, most samples in A2D dataset contain significantly more negative classes than positive
classes, thus causing severely skewed class distribution. My solution is to dynamically compute
and apply a ``pos-weight'' that has a length of the number of classes to all the classes in every
mini batch. For example, if there are 1 positive sample and 3 negative samples of a particular
class, ``pos-weight'' would be 3/1 = 3 for this class \cite{bce}. Then the loss function would act as if there
are balanced number of positive and negative samples in this class.

\section{Result}
The model did not converge well on training. Although training loss dropped significantly during the first epoch, it fluctuated approximately within a range of 0.1 to 0.25 starting from epoch 5.

On validation set, the model achieved the following performance:
\begin{itemize}
    \item Precision: 32.5
    \item Recall: 43.3
    \item F1-score: 34.6
\end{itemize}

\section{Conclusion}

In this report, I have developed a model from a backbone ResNet-152 model for performing multi-label actor-action classification. My method of adding class weights to every mini batch in training has reduced the effect of imbalanced data and thus improved validation accuracy. I have demonstrated that this can provide reasonably accurate recognition of the actor-action pairs given an image from the A2D dataset. F1-score achieved is 34.6.

{\small
	\putbib[students/kguo/egbib]
}
\end{bibunit}

\clearpage
\newpage

\setcounter{section}{0}
\setcounter{figure}{0}
\setcounter{equation}{0}
\setcounter{table}{0}

\title{Tri-Layer Convolutional Neural Network Model for Multi-Label Actor-Action Classification}

\author{
  Richard Magnotti \hspace{4mm} Andrew Sexton \hspace{4mm} Jeet Ketan Thaker\\
  University of Rochester\\
  {\tt\small \{rmagnott, asexton2, jthaker\}@ur.rochester.edu}
}

\maketitle

\begin{abstract}
    The goal of this project was to develop a novel neural network design to work on the A2D dataset. This dataset contains 3782 videos from YouTube which are annotated with an actor-action label. The primary directive was to build a model to predict the classes of actors and actions in each frame, some of which contain multiple actors or actions. The capability of the network is evaluated by the precision, recall and F1score of the network on the trained model. 
\end{abstract}

\begin{bibunit}[unsrt]
\section{Introduction}
Action recognition can take on a number of different forms, including human limb movements as well as the goals of the actions associated with them, such as exercising or clapping one’s hands \cite{DBLP:journals/corr/abs-1806-11230}. This research aims to explore a novel approach to action recognition by parallelizing multi-layer actor-action classifiers \cite{DBLP:journals/corr/XuXC17}, that in concept require multiple CNNs in sequence to pass data across them. We first make several attempts to construct various types of CNNs in PyTorch, and ultimately utilize built-in models to achieve the desired results. 

\begin{figure*}[t]
 \center
  \includegraphics[width=\textwidth]{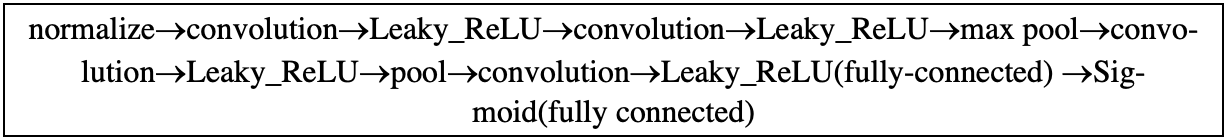}
  \caption{Vanilla CNN, first attempt.}
  \label{Figure 1}
\end{figure*}

\section{Methods}
It was decided to independently design two models simultaneously and then attempt to merge the best ideas from both. This was done in an effort to parallelize the work being done in order to take advantage of hardware opportunities, as well to develop a working model in light of local team meetings not being viable.  
 
Of note, all these models were tested with a variety of loss functions and optimizers. The loss functions included Mean Square Error (MSE), Multilabel Margin Loss, Cross Entropy Loss, and Binary Cross Entropy Loss (BCEWithLogitsLoss). Additionally, several different optimizers were utilized respectively, including Adam, AdamW and SGD. The first few models were built from the ground up using built in PyTorch packages. The final couple models were developed using pretrained PyTorch models. 
 
\section{Experiments}
\subsection{Handcrafted CNNs}
The first model was initially developed as a simple, “vanilla” four-layer CNN. For novelty and expected better loss, instead of a normal Leaky ReLU function we chose to use the “relu” as an activation function from the PyTorch library. The layers consisted of convolution, Leaky ReLU, max pooling, and finally the fully connected layers for prediction. This model performed very poorly, with all evaluation metrics below 5 (where the goal was ~30 for each category). Additionally, the final cost function was sigmoid. The generic structure can be seen in figure 1. 
 
The second attempt was done slightly differently. We hypothesized that adding more layers to a CNN would result in richer (and optimistically more accurate) detected features. Thus, we added more layers and included a Softmax function at the final output layer. This also performed very poorly with all metrics below a score of 5. 
 
The third attempt was dubbed “novel semi-fully connected CNN” by the team. We hypothesized that if we added a fully connected layer in the middle of the network, that it would make considerable improvements to the accuracy as a consequence of each neuron in the feature map being considered, as opposed to only part of it at a time as with regular convolutional layers. The structure is outline in figure 2. However, this did not change the evaluation metrics appreciably with all final metrics being below five. 
 
The last attempt for this model was to use the built-in PyTorch models, such as ResNet152, AlexNet, and VGG16. The pretrain flag was set to False, so that the structure of the model was used and not the already trained version. ResNet152, AlexNet, and VGG16 all performed very poorly on the metrics when trained on our data, and unfortunately on par with the handmade version listed above. This was not ideal, so at this point the focus was shifted to our second, pretrained, version which was having better results. 
 
The second model started off similarly, but backwards. 
The first attempt was using a simple seven-layer CNN  

in order to find a baseline. This model utilized Sigmoids, Tanh functions, Leaky ReLU as well as softmax. This also return very poor results, so the model was reduced to three layers, which also produced very poor results. 
 
As a final attempt at a handcrafted model, we attempted to design a “Tri Layer Model”, which consists of a classifier for each the actions, the actors and the actor-action pair and then to combine the three to produce a final output. All classifiers were seven-layer CNN models like the first attempt, and the combined layers were a fully connected network. This layer had 16 inputs, 7 for the actors and 9 for the actions, and 43 outputs for each of the actor-action pairs. This model performed poorly, much like its predecessors.  

\begin{figure*}[t]
 \center
  \includegraphics[width=\textwidth]{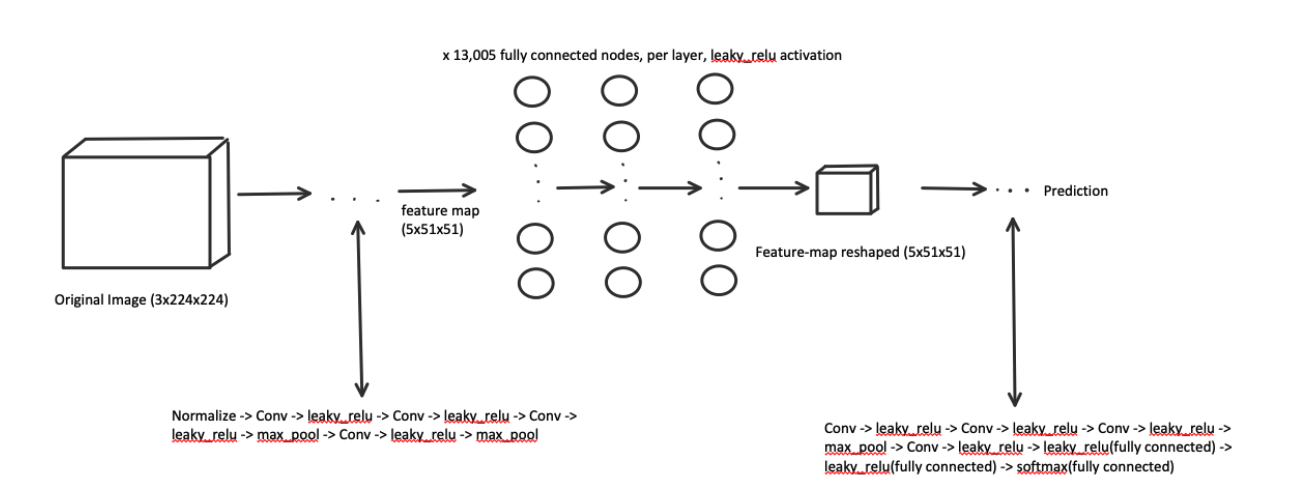}
  \caption{Semi-Fully-Connected ConvNet Structure.}
  \label{Figure 2}
\end{figure*}

\subsection{Pretrained CNNs}
It was then decided to utilize pre-trained models from the PyTorch library with the pretrained flag set to True. Our attempts at pretrained CNNs is outlined below.  
This time, the models used were ResNet18 version. The final output layer was set to 43 for each actor-action pair, and the first linear layer was trained, but the rest were left alone. This was the first attempt that actually performed up to the baseline requirements for the assignment. Utilizing binary cross entropy loss and the Adam optimizer, the evaluation metric scores were: Precision = 25.1, Recall = 36.7, F1 = 28.0. Our final attempt at a novel approach to CNN, was the TriLayer model. There were 3 classifiers each classifying actor, action, actor-action respectively. These results would then be combined and fed into a fully connected layer to output the required 43-dimensional label. Each classifier will use a pre-trained Resnet only training the last FC layer. This is done to preserve the filters Resnet already has learned. Different losses and optimizers were tried, and we selected BCEWithLogitsLoss as out loss function as it is widely used for multilabel classification and Adam optimizer with learning rate set to 0.0005 for 10 epochs and then 0.00003 for 10 more epochs.  
 
We concatenate the results from the 3 classifiers as shown in the diagram below and feed it into a fully connected layer. The hope is that it learns what the true labels are when given information about actors and actions.  

\begin{figure*}[t]
 \center
  \includegraphics[width=\textwidth]{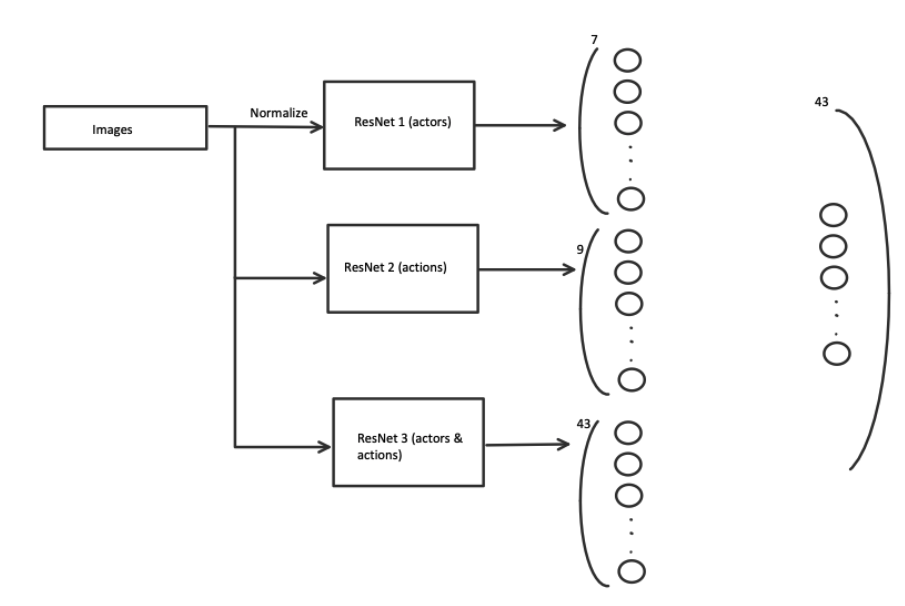}
  \caption{Pretrained model, Tri-Layer design.}
  \label{Figure 3}
\end{figure*}
 
\subsubsection{Tri Layer Model Overview Label Construction}
The labels provided are 43 dimensional vectors with the following schema:  
 
1.	adult-climbing  
2.	adult-crawling  
... 
43. dog - none  
 
We need to extract labels corresponding solely to actors and actions. There are 7 actors and 9 actions we need to consider.  
 
Label1 (for actors) = [adult, baby, ball, bird, car, cat, dog] 
Label2 (for actions) = [climbing, crawling, eating, jumping, rolling, running, walking, none, flying] 
 
A one hot encoding schema was followed to encode actors and actions. Therefore, for a single image we had two target vectors. One for actors and one for actions. 

\subsubsection{Tri-Layer Model Results}
The final result of this model when tested on the validation set is: Precision: 15.6, Recall: 36.7, F1: 19.1. This is below the baseline given to us, the reason we think that is, is because of the lack of data. The model makes sense on paper extracting specific information from each image and combining to give a desired output but in practice it did not work as well.  
 
\section{Conclusion}
It turns out to be very difficult to achieve performance increases over the models designed and packaged with the PyTorch library. With more access to resources such as GPU computing and perhaps without having a pandemic keeping everybody separated, more better designs could have been attempted. Novelty is a difficult concept to implement, because often the results don’t function at all and if they do, they do so poorly. The team was very interested in the “Tri Layer Model” approach, as it seemed quite novel indeed. However, although the network worked and produced outputs, the performance was generally poor, and we were not able to find an implementation of it that improved upon the initial scores. Ultimately, however, we were able to achieve the desired scores with a vanilla pretrained model and achieve reasonably close scores with a Tri-Layer approach utilizing pretrained models. 

{\small
	\putbib[students/richard/mybib]
}
\end{bibunit}

\newpage

\setcounter{section}{0}
\setcounter{figure}{0}
\setcounter{equation}{0}
\setcounter{table}{0}

\title{Incorporating Visual and Temporal Features for Actor-Action Classification}

\author{Yiyang Su\\
University of Rochester\\
{\tt\small ysu16@u.rochester.edu}
}

\maketitle

\begin{abstract}
   In this report, we attempt to presents an approach to the Actor-Action Classification problem that incorporates both visual and temporal information.
\end{abstract}
\begin{bibunit}[unsrt]
\section{Introduction}

The Actor-Action problem is the problem of determining the actor and the actor's action given a video. A video may have more than one actor-action pair.

In this project we use a variant of the A2D dataset (which can be found at \url{https://web.eecs.umich.edu/~jjcorso/r/a2d/}). The only difference between the dataset we use and orignal A2D dataset is that the dataset we used has been already divided into training, evaluation, and testing sets.

\section{Methodology}

\subsection{Network Architecture}

Our model consists of a combination of a Convolutional Neural Network (CNN) and Recurrent Neural Network (RNN), each followed by a linear layer.

More specifically, our network consists of the following layers:
\begin{itemize}
    \item Pretrained Resnet-152 \cite{he2016deep} without the last linear layer (i.e., Resnet-151), producing a 2048-dimensional feature.
    \item A linear layer with input dimension 2048 and output dimension 1024.
    \item A 2-layer LSTM with input dimension 2048 and hidden dimension 512.
    \item Three linear layers where the first layer has input dimension 3072 (for more details for why this is the case, please refer to Section 2.3) and the last layer has output dimension 43 with Sigmoid as their activation functions. Each linear layer has a smaller size than their previous layers.
\end{itemize}

We designed our model with the hope that in this way, pretrained CNN can extract visual features from the frames and LSTM can extract temporal features from other frames in the video. And thus, by combining these two models, our model can make use of both the visual features and temporal features.

    \subsection{The Modified Dataloader}

    To exploit the temporal features, it is necessary to modify the dataloader so that it does more than returning one frame each time.

    One intuitive approach would be to return all previous frames available in the dataset of the same video. But as we have experimented, this consumes all available memory resources on our training machine in one step.

    We have adopted a less memory-demanding approach, which is to load 5 previous frames that are closest to the target frame. In order to support SGD with batch size greater than 1, we padded zero frames in the front for frames that do not have five previous frames available in the data set. Therefore, each sample from the data set is a Tensor of size $6\times 3 \times 224 \times 224$.

   \subsection{Putting it together}

    To summarize, our approach is to pass each of the 6 frames through the CNN, obtain a 2048 dimensional feature for each frame and then treat them as a sequence of frames with length 6 and pass them through the LSTM and get output of size $6 \times 1 \times 512$. And finally, we flatten it, pass the resultant 3072 dimensional vector through the linear layers, and get a 43 dimensional vector, which is the classification result.

    In training the model, we use a Adam optimizer \cite{kingma2014adam} with an initial learning rate of $0.001$. We use a batch size of 4, a binary cross entropy loss (\verb|torch.nn.BCELoss()|), and trained the model for three epochs.

\section{Conclusion}

In this report, we present a potential approach to incorporate both visual and temporal features for the Actor-Action Classification problem. However, more work has to done in order to evaluate the performance of this approach and to verify whether this approach is superior to using solely visual features (it is unlikely that a model uses solely temporal features because temporal features depend on visual features).

{\small
\putbib[students/ysu16/egbib]
}

\end{bibunit}
\clearpage

\end{document}